\documentclass[11pt]{article}

\usepackage[final]{acl}

\usepackage{times}
\usepackage{latexsym}

\usepackage[T1]{fontenc}

\usepackage[utf8]{inputenc}

\usepackage{microtype}

\usepackage{inconsolata}

\usepackage{graphicx}

\usepackage{url}
\usepackage{tcolorbox}
\usepackage{wrapfig}
\usepackage[utf8]{inputenc} 
\usepackage[T1]{fontenc}    
\usepackage{hyperref}       
\usepackage{url}            
\usepackage{booktabs}       
\usepackage{amsfonts}       
\usepackage{nicefrac}       
\usepackage{microtype}      
\usepackage{xcolor}         
\usepackage{amsmath,amssymb}
\usepackage{algorithmic}
\usepackage{multirow}
\usepackage{xcolor}
\usepackage{tcolorbox}
\tcbuselibrary{breakable}
\usepackage{paracol}
\title{SelfBudgeter: Adaptive Token Allocation \\ for Efficient LLM Reasoning}

\author{%
Zheng Li\textsuperscript{1}, Qingxiu Dong\textsuperscript{1}, Jingyuan Ma\textsuperscript{1}, Di Zhang\textsuperscript{1}, Kai Jia\textsuperscript{2}, Zhifang Sui\textsuperscript{1} \\ 
\textsuperscript{1}State Key Laboratory for Multimedia Information Processing, \\
School of Computer Science, Peking University\\ 
\textsuperscript{2}BandAI, Bytedance\\
\texttt{\{lizheng2001,szf\}@pku.edu.cn} \\ 
\texttt{\{dqx,mjy,zhangdi\}@stu.pku.edu.cn} \\ 
\texttt{jiakai@bytedance.com} \\ 
}

\begin{document}
\maketitle
\begin{abstract}
Recently, large reasoning models demonstrate exceptional performance on various tasks. 
However, reasoning models always consume excessive tokens even for simple queries, leading to resource waste and prolonged user latency.
To address this challenge, we propose SelfBudgeter - a self-adaptive reasoning strategy for efficient and controllable reasoning.
Specifically, we first train the model to self-estimate the required reasoning budget based on the query. We then introduce budget-guided GRPO for reinforcement learning, which effectively maintains accuracy while reducing output length.
Experimental results demonstrate that SelfBudgeter dynamically allocates budgets according to problem complexity, achieving an average response length compression of 61\% on math reasoning tasks while maintaining accuracy.
Furthermore, SelfBudgeter allows users to see how long generation will take and decide whether to continue or stop. Additionally, users can directly control the reasoning length by setting token budgets upfront.
\end{abstract}

\section{Introduction}
\label{introduction}
Recent large reasoning models, such as O1~\citep{openaiO1}, have shown remarkable performance in various complex reasoning tasks~\citep{deepseekr1,qwq-32b-preview}.
The primary success factor lies in the long chain of thought (CoT) process learned through reinforcement learning (RL), which  allows the model to break down reasoning steps and scaling test-time compute~\citep{test-time-compute,deepscaler2025}. 



However, reasoning models tend to use overly long thought processes even for simple questions. This ``overthinking'' phenomenon leads to a waste of computational resources and excessive user waiting times~\citep{chen2024think23overthinkingo1like,sui2025stop}. 
For example, when answering simple questions such as ``What is the answer to 2+3?'', the QwQ-32B model provides 13 different solutions and generates 100 times more tokens than Qwen2.5-72B-Instruct model~\citep{qwen2025qwen25technicalreport}.

Prior studies have explored various approaches to mitigate overthinking through response length control and computation routing. Existing methods mainly include: (1) Prompt-based approaches~\citep{DBLP:tokencomplexity,DBLP:cod} that implicitly guide length through instructions, (2) Integrated training strategies that teach models to adaptively determine reasoning steps via SFT~\citep{DBLP:self-training,DBLP:cot-valve} or RL with length penalties~\citep{aggarwal2025l1,DBLP:trainingllmtoreasonefficiently}, and (3) Router-based~\citep{aytes2025sketch,chuang2025confident} architectures employing classifiers to allocate computation paths. While achieving partial progress, these methods either lack precise length control, require additional computational overhead, or fail to explicitly output optimal reasoning lengths~\citep{aggarwal2025l1,xu2025scalablechainthoughtselastic}.
\begin{figure}[htbp]
\centering
\includegraphics[width=\linewidth]{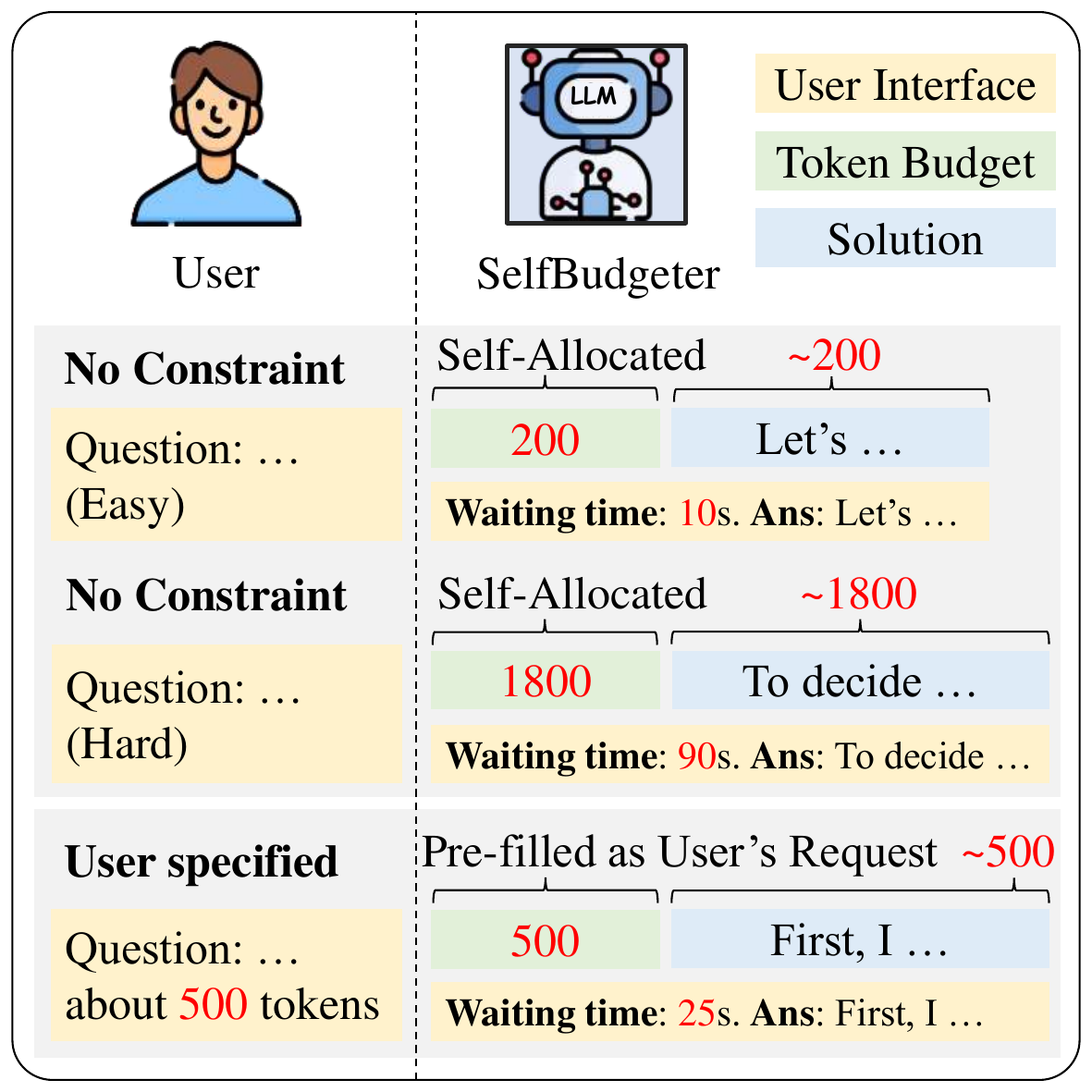}
\caption{Overview of the SelfBudgeter. SelfBudgeter's responses comprise two sections: \textbf{Token Budget} and \textbf{Solution}. For unrestricted queries, it estimates tokens needed for the \textbf{Solution} based on problem complexity. When users specify requirements, it pre-fills the \textbf{Token Budget} accordingly. The \textbf{Solution} generation strictly adheres to \textbf{Token Budget} limits, whose numerical value indicates anticipated response duration.
 }
\label{fig:teaser}
\end{figure}

We propose \textbf{SelfBudgeter} that enables reasoning models to (1) estimate the minimal token budget required for correct responses when users do not specify token constraints, and (2) generate responses of corresponding lengths while adhering to either self-estimated or user-defined token budgets. SelfBudgeter aims to mitigate the overthinking issue by predicting the minimal possible token 
budget, thereby significantly reducing user waiting time. As shown in Figure~\ref{fig:teaser}, SelfBudgeter can provide a relatively accurate token budget estimation before generating responses, so users can precisely anticipate the waiting time and decide whether to wait for the full output or terminate early based on their needs. Additionally, when specific requirements arise, users can pre-fill the token budget field to constrain the model’s response within the given limit, thereby improving interaction efficiency.

Our training framework consists of two main stages. During the Cold-Start stage, the model learns to predict a token budget using \texttt{<budget>} tags. Subsequently, in the RL training stage, we optimize SelfBudgeter to improve accuracy while producing concise and budget-consistent outputs. Our main contributions are as follows: 
\begin{itemize}
    \item We propose \textbf{SelfBudgeter}, a training framework that enables large language models to explicitly predict and control their output length via a self-estimated token budget.
    \item We demonstrate that SelfBudgeter can significantly reduce response length while preserving model performance, and in some cases even improving accuracy, showing that length control does not necessarily degrade reasoning quality.
    \item We validate the scalability of SelfBudgeter across different model sizes, with consistent benefits observed when applying the framework to larger models.
\end{itemize}

\section{Related Work}
\label{relatedwork}

\paragraph{Overthinking in LLMs}

The emergence of the reasoning models such as O1, DeepSeek-R1 and QwQ has advanced complex problem-solving through longer CoT~\citep{Wei2022ChainOT, openaiO1, deepseekr1, qwq-32b-preview}. 
However, researchers observed a tendency for reasoning models to overthink simple problems, incurring unnecessary computational cost and even degrading accuracy~\citep{chen2024think23overthinkingo1like,sui2025stop,DBLP:whenmoreisless,zeng-etal-2025-revisiting}. 
Current solutions for overthinking mainly fall into three categories. Prompt-based methods try to control response length by adding instructions in prompts, but lack precise control~\citep{DBLP:tokencomplexity,DBLP:ccot,DBLP:cod,conciseThoughts}. 
Integrated Training-based methods, including supervised fine-tuning with variable-length data and reinforcement learning with length-aware rewards, encourage adaptive reasoning depth, yet cannot reliably satisfy user-specified length constraints~\citep{DBLP:self-training,DBLP:cot-valve,DBLP:skipsteps,han2024token,DBLP:c3ot,DBLP:tokenskip,yang2025towards,aggarwal2025l1,DBLP:trainingllmtoreasonefficiently,DBLP:o1-pruner,chen2025towards,DBLP:demystifying,xu2025scalablechainthoughtselastic,yang2025dynamicearlyexitreasoning}. 
Router-based methods introduce an auxiliary classifier to dispatch queries to different models, but incur additional computational overhead~\citep{aytes2025sketch,chuang2025confident,DBLP:self-ref,DBLP:routellm,pan2025specreasonfastaccurateinferencetime}. 
Current methods either sacrifice precise control, require extra computation, or fail to bridge autonomous budget estimation with strict adherence.

\paragraph{Token Budget}
In addressing the issue of overthinking, a highly intuitive approach involves directly constraining the output length. CCoT~\citep{conciseThoughts} attempts to achieve this by incorporating a word budget into the prompt, various approaches—including character, token, and step budgets~\citep{DBLP:tokencomplexity}—have been attempted by directly incorporating them into prompts, yet achieving precise control over the model's output behavior remains challenging. TALE~\citep{han2024token} introduces, for the first time, the concept of a token budget. 
TOPS~\citep{yang2025towards} attempts to enable the model to autonomously determine the required effort for solving a given task. However, both TALE and TOPS fail to explicitly guide the model to produce the optimal token budget. They also fail to effectively control the output length according to a given token budget. L1~\citep{aggarwal2025l1} and Elastic Reasoning~\citep{xu2025scalablechainthoughtselastic} can more precisely control the output length under a given token budget, yet they fail to enable the model to autonomously estimate an appropriate response length. Our proposed method enables the model to autonomously estimate the optimal token budget and subsequently generate text in strict adherence to it.

\section{Method}
\label{method}

To minimize the overthinking problem in LLMs, 
we propose SelfBudgeter for efficient reasoning.
Our method aims to enable the model to autonomously determine an appropriate token budget and generate responses of corresponding length while adhering to this budget. 
Although reasoning models may occasionally overthink simple problems, their response lengths generally increase with problem difficulty. This phenomenon demonstrates that the model possesses the capability to allocate token quantities reasonably based on problem complexity. Previous works such as L1~\citep{aggarwal2025l1} and Elastic Reasoning~\citep{xu2025scalablechainthoughtselastic} have also demonstrated that models can generate responses of appropriate length according to a given token budget.

Therefore, we design SelfBudgeter, which employs a reward function to guide the model in: (1) learning an output format where it first predicts a token budget before generating the answer, (2) allocating appropriate token budgets based on its own capabilities and question difficulty, and (3) generating solutions with optimal length while ensuring answer accuracy.

\subsection{SelfBudgeter}
SelfBudgeter is a concise and efficient method for automatic and precise length control. We design the Precise Budget Control Reward (PreB Reward) to achieve precise control over length. The detailed introduction of PreB Reward can be found in Section~\ref{method:preb reward}.
We employ the GRPO algorithm to train the model in predicting appropriate token budgets based on problem difficulty and generating responses with lengths conforming to the specified budget.

Our reward function is formally defined as Formula \ref{formula:reward function}:
\begin{equation}
\small
\begin{aligned}
R(C,F,\ell,b,b_{\max}) &= \\[2pt]
&\hspace{-2.4cm}   
\left\{
\begin{array}{l}
r_f, \\
\qquad\qquad\qquad\qquad\qquad\qquad\text{if } F = 0, \\[6pt]
\text{P}_\text{B}(b,b_{\max})
+ \text{PreB}\!\big(
    s_{\min}^w,\, s_{\max}^w,\,
    \ell,\, b,\, \alpha,\, b_{\text{best}}^w
\big), \\
\qquad\qquad\qquad\qquad\qquad\qquad\text{if } F = 1 \text{ and } C = 0, \\[6pt]
\text{P}_\text{B}(b,b_{\max})
+ \text{PreB}\!\big(
    s_{\min}^c,\, s_{\max}^c,\,
    \ell,\, b,\, \alpha,\, b_{\text{best}}^c
\big), \\
\qquad\qquad\qquad\qquad\qquad\qquad\text{if } F = 1 \text{ and } C = 1.
\end{array}
\right.
\end{aligned}
\label{formula:reward function}
\end{equation}

where
\begin{equation}
b_{\text{best}}^c=(1-\alpha)\cdot b,\quad b_{\text{best}}^w=(1+\alpha)\cdot b
\label{formula:b best}
\end{equation}
\[
\begin{aligned}
&\textbf{Hyperparameters:} \\[-2pt]
&\begin{aligned}
   &\quad r_f              &:\ &\text{Penalty for format error}, &\\[-2pt]
   &\quad s_{\min}^{w/c}   &:\ &\text{Minimum reward (wrong/correct)}, &\\[-2pt]
   &\quad s_{\max}^{w/c}   &:\ &\text{Maximum reward (wrong/correct)}, &\\[-2pt]
   &\quad \alpha           &:\ &\text{Tightness coefficient of budget}, &\\[-2pt]
   &\quad r_b              &:\ &\text{Penalty for excessive budget},&
\end{aligned}
\\[4pt]
&\textbf{Inputs:} \\[-2pt]
&\begin{aligned}
   &\quad C         &:\ &\text{Correctness for answer}, &\\[-2pt]
   &\quad F         &:\ &\text{Correctness for format}, &\\[-2pt]
   &\quad l         &:\ &\text{Response length}, &\\[-2pt]
   &\quad b         &:\ &\text{Model's budget}, &\\[-2pt]
   &\quad b_{\max}  &:\ &\text{Maximum acceptable budget}.&
\end{aligned}
\end{aligned}
\]

To ensure stable prediction of the token budget prior to response generation, any responses deviating from the prescribed format will be assigned the minimum reward score of $r_f$. For responses that adhere to the required format, we further encourage efficient reasoning through two complementary reward components. The Budget Penalty ($\text{P}_\text{B}$) incentivizes the model to continuously reduce its predicted budget, while the Precise Budget Control Reward ($\text{PreB}$) aligns the actual response length with the model’s self-predicted budget. Together, these mechanisms enable the model to both forecast an appropriate budget and generate shorter, budget-conformant outputs.


\subsection{Budget Penalty}
To enable the model to learn token budget allocation, we introduce a budget penalty module defined by Formula \ref{formula:budget penalty}. The model incurs a penalty $r_b$ when its estimated token budget exceeds the maximum acceptable budget $b_\text{max}$. No penalty is applied when the estimated token budget remains within $b_\text{max}$. A detailed introduction of $b_\text{max}$ is presented in Section~\ref{method:data preprocessing}. Briefly stated, for a given question, $b_\text{max}$ equals the response length if the base model can answer it correctly; otherwise, $b_\text{max}$ is set to $\infty$.
\begin{equation}
\text{P}_\text{B}(b, b_\text{max}) = 
\begin{cases} 
0, & \text{if } b \leq b_\text{max}, \\ 
r_b, & \text{else.}
\end{cases}
\label{formula:budget penalty}
\end{equation}

\subsection{Precise Budget Control Reward}
\label{method:preb reward}


Inspired by the cosine reward~\citep{DBLP:demystifying}, we propose the Precise Budget Control Reward (PreB Reward). While the cosine reward helps mitigate overthinking tendencies, it lacks precise control over output length, as it only constrains the upper bound of the response. To address this limitation, we introduce a tightness coefficient $\alpha$ to better align the response length with the specified token budget.

Given the inherent challenge for models to precisely comply with token budgets, we relax the length constraint to require only approximate adherence within $\alpha \cdot b$ around the target budget $b$. As shown in Formula \ref{formula:cosine reward}, when the model's response length falls outside the specified range, the corresponding reward score plummets to its minimum value $s_\text{min}$.

For incorrect responses, the function incentivizes longer reasoning chains (increasing length $\ell$) to encourage deeper analysis that might lead to correct conclusions. Conversely, for correct answers, the reward peaks at the minimally sufficient length $(1-\alpha)\cdot b$ to prevent unnecessary computational overhead while maintaining accuracy. This explains why in Formula \ref{formula:b best}, the value of $b_\text{best}$ differs between correct and incorrect responses from the model. This dual mechanism promotes efficient reasoning by adaptively modulating response lengths based on answer correctness.
\begin{equation}
\small
\begin{aligned}
\mathrm{PreB}(s_{\min}, s_{\max}, \ell, b, \alpha, b_{\text{best}}) &= \\[4pt]
&\hspace{-4.0cm}
\left\{
\begin{array}{l}
s_{\min},\qquad\qquad\qquad\qquad\ \ \ \text{if } \dfrac{|\ell - b|}{b} > \alpha, \\[10pt]
s_{\min} + (s_{\max} - s_{\min}) \times \\[2pt]
\dfrac{1}{2}\!\left( 1 + 
\cos\!\left( 
\pi \cdot \dfrac{|\ell - b_{\mathrm{best}}|}{2\alpha b}
\right)
\right),\qquad\text{else.}
\end{array}
\right.
\end{aligned}
\label{formula:cosine reward}
\end{equation}

\subsection{Accuracy Reward}
To ensure the model's post-training accuracy does not degrade below its initial performance, we configure hyperparameters to guarantee that the minimum reward for correct responses always exceeds the maximum reward for incorrect responses. Specifically, our design ensures that: A correct response, which has a token budget exceeding $b_\text{max}$ and receives the lowest budget following reward $s^C_\text{min}$, will yield a higher total reward than an incorrect response that has a token budget within $b_\text{max}$ and receives the highest budget following reward $s^W_\text{max}$. This constraint is formally expressed as: $s^C_\text{min}+r_b \ge s^W_\text{max}$.

Overall, the core design of SelfBudgeter consists of three key modules: Budget Penalty, PreB Reward, and Accuracy Reward, which collectively balance length compression, correctness, and precise length control--ultimately delivering a better user experience.

\section{Experiment}
\label{experiment}
\begin{figure*}[htbp]
    \centering
    \includegraphics[width=\textwidth]{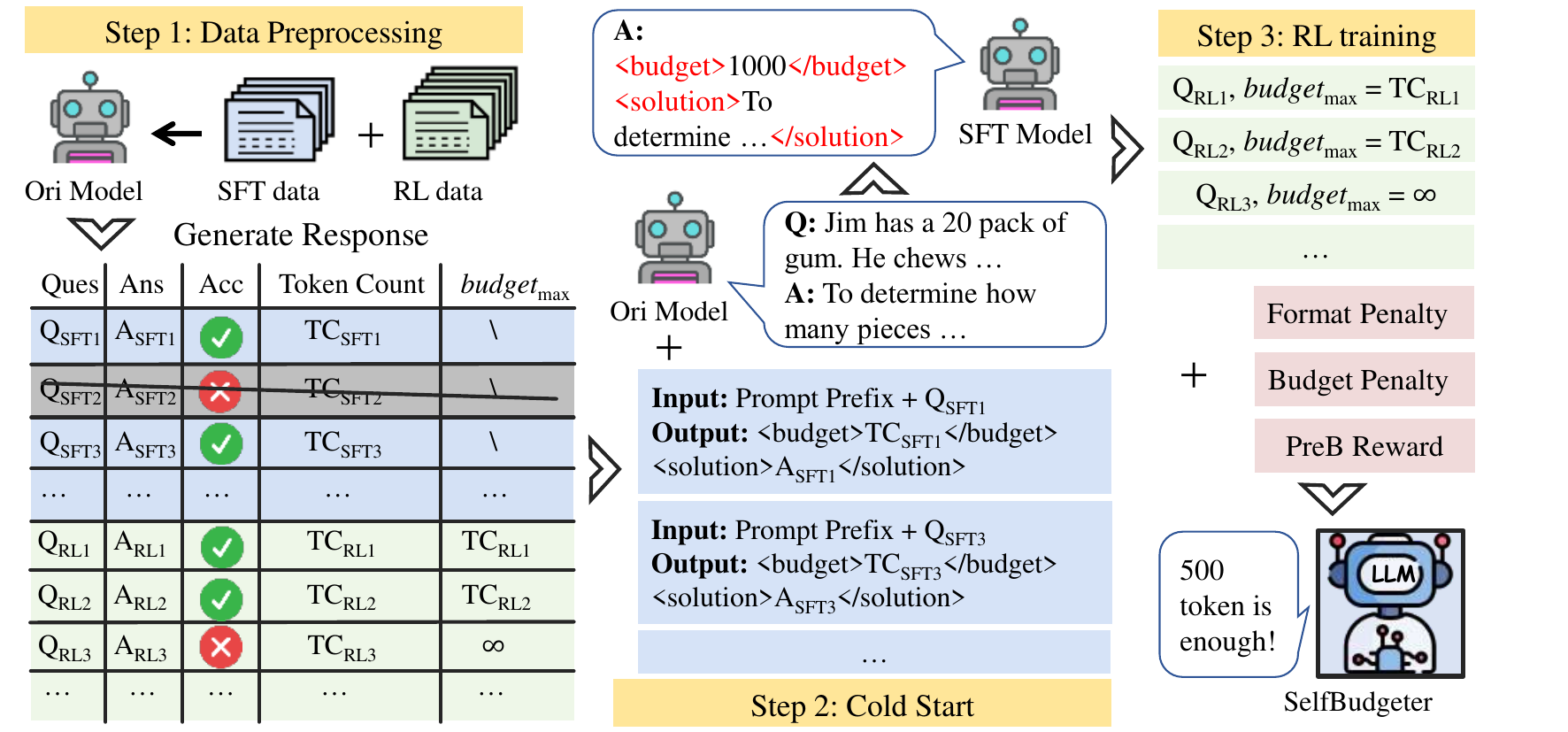}
    \caption{Overview of the SelfBudgeter training framework. The workflow consists of three key steps: (1) \textbf{Data preprocessing}: The initial model preprocesses the data to compute token budgets for subsequent training; (2) \textbf{Cold-start}: The model is fine-tuned to adopt the new output format; (3) \textbf{RL Training}: Through reward functions, the model learns to allocate token budgets and generate compliant outputs.}
    \label{fig:autobudgeter}
\end{figure*}

\subsection{Training Template}

The existing reasoning models utilize a pair of \texttt{<think></think>} tags to demarcate the thinking process from the final solution output. Building upon this format, we have further incorporated a token budget component.

To enable the model to dynamically allocate token usage based on question difficulty, we design an output format as follows:

\begin{verbatim}
<budget>an integer</budget><solution>
response</solution>
\end{verbatim}

The format requires the model to first estimate the required token budget before providing the answer to the question. When no user constraint exists, the model autonomously predicts the token budget. When users specify a token limit, we pre-fill the <budget> field and let the model generate the <solution> within this constraint.

\subsection{Data Preprocessing}
\label{method:data preprocessing}


At this stage, we collect the model's responses to the test questions used in both the cold-start and RL training phases, and then evaluate the correctness and length of the responses.  

For the cold-start data, we retain the model's correct responses along with their lengths and discard incorrect answers to prevent reinforcing the model's memory of wrong responses.  

For the RL training data, we calculate $budget_
\text{max}$ (for convenience, we will refer to it as $b_\text{max}$ in the following sections) using Formula \ref{formula:b_max}, representing the maximum acceptable token budget for a given question. When the model answers correctly, the correctness of the response indicates that the minimum token budget required for a correct answer does not exceed the current length. Therefore, we encourage the model to further compress the response length and set $b_\text{max}$ to the current response length. When the model answers incorrectly, the relationship between the minimum token budget needed for correctness and the current length remains unclear, so any token budget is acceptable.
\begin{equation}
b_\text{max} =
\begin{cases} 
\text{response length}, & \text{if C=1}, \\ 
{\infty}, & \text{else.}
\end{cases}
\label{formula:b_max}
\end{equation}

\subsection{Cold Start}
\label{method:cold start}
In our actual RL training process, we observe that requiring the model to simultaneously master multiple objectives - learning the new output format, providing appropriate token budgets, generating solutions of corresponding lengths according to the budget, while maintaining or improving accuracy - proved excessively challenging. After extended training periods, the model often only succeeds in adopting the output format without achieving the other goals. Inspired by the DeepSeek-R1 training methodology, we introduce a cold-start phase to accelerate training and enable the model to first learn the new output format before proceeding to more complex tasks. The overall training framework is illustrated in Figure~\ref{fig:autobudgeter}.

To prevent the model from losing its original reasoning capability during the cold-start phase, fine-tuning must be performed using either the model's own generated responses or datasets containing long CoT responses. In our approach, we pre-populate the \texttt{<budget>} section with token counts obtained during the preprocessing stage. The \texttt{<solution>} section is filled with the model's generated responses. The instruction prefix we prepend to each question can be found in Appendix~\ref{appendix:cold-start}.

\begin{table*}[t]
\centering
\begin{tabular}{lrrrrrr}
\toprule
\multirow{2}{*}{Models} & \multicolumn{2}{c}{GSM8K} & \multicolumn{2}{c}{MATH500} & \multicolumn{2}{c}{AIME2025} \\
\cmidrule(lr){2-3} \cmidrule(lr){4-5} \cmidrule(lr){6-7}
 & Acc & Len & Acc & Len & Acc & Len \\
\midrule
DeepSeek-R1-Distill-Qwen-1.5B & 73.09  & 2865.08  & 74.93  & 5327.12  & \textbf{22.22}  & 14444.03  \\
E1-Math-1.5B(0.5K,1K) & 60.20  & 1205.21  & 35.53  & 1499.54  & 4.44  & 3008.44  \\
E1-Math-1.5B(4K,1K) & 72.10  & 1299.62  & 72.47  & 2088.44  & \underline{21.11}  & 5578.13  \\
L1-Max(3600) & \underline{79.56}  & 571.72  & \underline{76.73}  & 1753.42  & 17.88  & 5213.89  \\
SelfBudgeter-1.5B & \textbf{84.10}  & 1231.79  & \textbf{78.47}  & 2326.85 & \underline{21.11} & 4288.10 \\
\midrule[\heavyrulewidth] 
DeepSeek-R1-Distill-Qwen-7B & 87.09 & 1918.21 & \underline{86.73} & 5387.19 & \underline{28.89} & 22158.79 \\
Eurus-2-7B-PRIME & \textbf{90.98} & 302.72 & 79.73  & 582.58 & 15.56  & 1254.52 \\
Qwen-2.5-7B-Simple-RL & 75.94 & 519.07 & 61.13 & 823.89 & 6.67 & 1429.94 \\
SelfBudgeter-7B & \underline{90.30} & 991.13 & \textbf{86.87}  & 2666.58 & \textbf{30.00} & 12241.84 \\
\bottomrule
\end{tabular}
\caption{Performance comparison on GSM8K, MATH500, and AIME2025. 
Accuracy (Acc) is reported in percentage, and length (Len) in tokens.}
\label{tab:main results performance}
\end{table*}
\subsection{Experiment Settings}
\label{model and datasets}
We conduct experiments on the DeepSeek-R1-Distill-Qwen-1.5B (R1-1.5B) model. We reproduce L1-Max using R1-1.5B, and select R1-1.5B and L1-Max as baseline methods for comparative evaluation against SelfBudgeter. In addition, we extend our experiments to the larger DeepSeek-R1-Distill-Qwen-7B (R1-7B) model. For more comprehensive comparison, we also include E1-Math-1.5B, R1-7B, Eurus-2-7B-PRIME~\citep{eurus7b}, and Qwen-2.5-7B-Simple-RL~\citep{simplerl7b} as additional baselines.



During the cold-start phase, we employ three datasets of varying difficulty—GSM8K~\citep{DBLP:gsm}, MATH~\citep{DBLP:math}, and s1k-1.1~\citep{DBLP:s1dataset}—to help the model learn the new output format while producing token budgets with diverse distributions. The s1k-1.1 dataset contains 1,000 challenging mathematical problems with long reasoning chains generated by DeepSeek-R1, which support both reasoning ability and format adaptation. For GSM8K and MATH, we select 1,500 training samples each that the model can answer correctly. For s1k-1.1, we directly use the native responses and compute the corresponding token counts with the model’s tokenizer to populate our designed template; in total, we retain 630 problems that DeepSeek-R1 answered correctly. This yields a training set of 3,630 samples. Following the preprocessing protocol in Sections~\ref{method:data preprocessing} and~\ref{method:cold start}, we fine-tune the model for one epoch. Throughout data collection and training, the model’s temperature is consistently set to 0.6.

During the reinforcement learning phase, we use STILL-3-Preview-RL-Data~\citep{Still3} dataset. It also serves as the training dataset for reproducing L1-max. This dataset collects 30K high-quality samples based on the MATH~\citep{DBLP:math}, NuminaMathCoT~\citep{numina_math_datasets}, and AIME 1983-2023~\citep{aime_1983_2024} datasets. It includes problems of varying difficulty levels, which also helps the model learn to allocate token counts adaptively based on difficulty. As described in Section \ref{method:data preprocessing}, we compute the maximum acceptable budget ($b_\text{max}$) based on the model's responses, then train the model for 3 epochs on this dataset. More detailed information can be found in Appendix~\ref{appendix:training details}.

\subsection{Main Results}

Table \ref{tab:main results performance} presents a comprehensive comparison of model performance on the GSM8K, MATH500, and AIME2025 test sets, evaluated in terms of accuracy (Acc) and average response length (Len). The table contrasts baseline models with different variants of the SelfBudgeter framework across varying model scales. For clarity, the best performance is highlighted in bold, while the second-best performance is indicated with \underline{underline}. It is worth noting that token limits for L1 are explicitly specified through prompt templates, whereas those for E1 are enforced via hard truncation. In contrast, SelfBudgeter autonomously estimates its token constraints during inference. All reported results are averaged over three runs with different random seeds.

\begin{figure*}[htbp]
    \centering
    \includegraphics[width=1\linewidth]{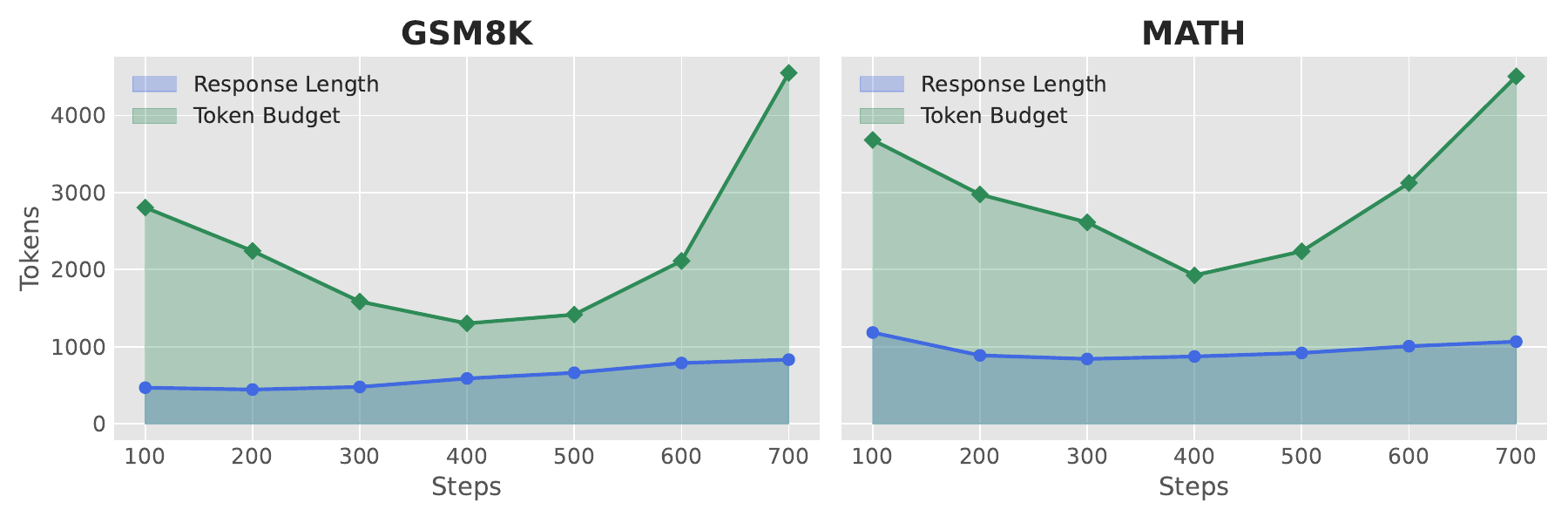}
    \caption{Response length and token budget on GSM8K and MATH benchmarks across training steps with $\alpha=0.5$. The curves show how the average response length (solid circles) and allocated token budget (solid diamonds) evolve during training.}
    \label{fig:dynamic alpha}
\end{figure*}

\paragraph{Baseline Comparison} Although the DeepSeek-R1-Distill-Qwen-1.5B baseline demonstrates strong accuracy, it requires substantially longer responses. On GSM8K, our method improves accuracy by 11.01 percentage points while compressing response length to 43\% of the original. On MATH500, it achieves a 3.54-point accuracy gain with response length reduced to 44\%. On AIME2025, our approach compresses response length to 30\% of the original while maintaining comparable accuracy. In contrast, although L1 and E1 attain stronger compression on certain datasets, they incur larger accuracy losses—L1 performs poorly on the challenging AIME2025 benchmark, while E1 suffers more pronounced accuracy degradation on the simpler GSM8K and MATH500 datasets.

In addition, Table~\ref{tab:main results performance} highlights that SelfBudgeter consistently strikes a better balance between accuracy and response length than existing baselines. Unlike L1, which enforces explicit length limits but collapses on AIME2025, or E1, which relies on hard truncation and severely harms accuracy, SelfBudgeter autonomously learns effective token budgeting. As a result, it achieves the best or second-best accuracy across all datasets while simultaneously reducing response length substantially.

Beyond its effectiveness at the 1.5B scale, our method also delivers efficient reasoning with larger models. SelfBudgeter-7B achieves the highest accuracy on MATH500 and AIME2025, and the second-best accuracy on GSM8K—only 0.68 points lower than the best-performing model. Meanwhile, SelfBudgeter-7B attains an average compression ratio of 48\%, further demonstrating the generality of our approach and its effectiveness at larger model scales. Compared with Eurus-2-7B-PRIME, which excels only on GSM8K but falls behind on harder reasoning tasks, and Qwen-2.5-7B-Simple-RL, which underperforms across all benchmarks, SelfBudgeter exhibits robust gains across datasets of varying difficulty.

\subsection{Dynamic Alpha Schedule}
In SelfBudgeter, $\alpha$ serves as a critical hyperparameter. As shown in Figure~\ref{fig:dynamic alpha}, we observe that using a fixed and relatively loose $\alpha$ can lead to \textit{reward hacking}: once the model learns to align the budget with the actual response length, it tends to inflate the predicted budget during later training stages, pushing the output length toward the lower bound of the acceptable range to obtain higher PreB scores. Conversely, when $\alpha$ is fixed but relatively tight, the token budget quickly collapses to the response length, which hinders the model from learning an optimal budgeting strategy. To address these issues, we introduce a \textit{dynamic alpha schedule}, where $\alpha$ is linearly decreased over training steps. This gradually tightens the tolerance range for acceptable response lengths and encourages closer convergence between the predicted budget and the actual output length. Consequently, the optimal $\alpha$ is not static but evolves throughout the training process.

Formally, the dynamic $\alpha$ is defined by a linear schedule:
\begin{equation}
\alpha_{\text{now}} = \alpha_{\text{start}} - \left(\alpha_{\text{start}} - \alpha_{\text{end}}\right) \cdot \frac{\text{step}_{\text{now}}}{\text{Total steps}}.
\end{equation}

This schedule only requires specifying the starting and ending values of $\alpha$ (i.e., $\alpha_{\text{start}}$ and $\alpha_{\text{end}}$), which are set to $6.0$ and $0.1$, respectively.

\section{Discussion}
\label{analysis}

\begin{table*}[htbp]
\centering
\small
\begin{tabular}{lrrrrrr}
\toprule
\multirow{2}{*}{Methods} & \multicolumn{2}{c}{GSM8K} & \multicolumn{2}{c}{MATH500} & \multicolumn{2}{c}{AIME2025} \\
\cmidrule(lr){2-3} \cmidrule(lr){4-5} \cmidrule(lr){6-7}
 & Acc & Len & Acc & Len & Acc & Len \\
\midrule
R1-1.5B &
73.09  & 2865.08  &
74.93  & 5327.12  &
22.22  & 14444.03 \\
Format Only &
1.64  & 4866.40  &
1.47  & 5558.35  &
0.00  & 6448.66 \\
F+Correctness &
84.58 & 1258.45  &
76.27 & 3123.37  &
20.00 & 11095.39 \\
F+C+PreB &
82.89 & 2151.26 &
78.00 & 3109.43 &
15.56 & 4671.86 \\
F+C+BudgetPenalty &
84.26 & 1263.38  &
76.67 & 2983.28  &
21.11 & 7263.38  \\
SelfBudgeter-1.5B &
84.10 & 1231.79 &
78.47 & 2326.85 &
21.11 & 4288.10 \\
\bottomrule
\end{tabular}
\caption{Ablation of Reward Components on Accuracy and Response Length}
\label{tab:ablation reward}
\end{table*}

\begin{table*}[htbp]
\centering
\begin{tabular}{lcccccc}
\toprule
$\alpha$ Scheduling &
\multicolumn{2}{c}{GSM8K} &
\multicolumn{2}{c}{Math500} &
\multicolumn{2}{c}{AIME2025} \\
\cmidrule(lr){2-3} \cmidrule(lr){4-5} \cmidrule(lr){6-7}
& Acc & Len
& Acc & Len
& Acc & Len \\
\midrule
Fixed Scheduling &
79.88  & 709.49 &
73.06  & 1015.96 &
10.00  & 1545.50 \\
Cosine Scheduling &
83.22  & 1149.43  &
75.93  & 1769.06  &
17.78  & 2366.34  \\
Linear Scheduling &
84.10  & 1231.79 &
78.47  & 2326.85 &
21.11  & 4288.10 \\
\bottomrule
\end{tabular}
\caption{Comparison of $\alpha$ scheduling strategies.}
\label{tab:alpha-scheduling}
\end{table*}

In the Discussion section, we further investigate the underlying mechanisms of SelfBudgeter from both optimization and behavioral perspectives. We first present an ablation study on reward components, analyzing how different reward terms contribute to length control and answer accuracy, and elucidating how SelfBudgeter achieves effective budget-aware reasoning. We then examine the impact of different $\alpha$ scheduling strategies, highlighting their influence on the final performance and stability of the learned budget allocation policy. In addition, we evaluate SelfBudgeter on general-domain knowledge reasoning benchmarks to assess its generalization ability beyond mathematical tasks, showing that it can retain competitive accuracy while substantially reducing reasoning length.

Beyond the main discussion, we provide additional analyses in the appendix to offer a more comprehensive understanding of SelfBudgeter’s behavior. Specifically, Appendix~\ref{appendix:prefill budget} studies the model’s adherence to pre-filled token budgets, Appendix~\ref{appendix: adaptive budget allocation} analyzes its ability to dynamically adjust budgets according to problem complexity, and Appendix~\ref{appendix:budget constrained generation} examines compliance with token constraints while preserving response quality. Finally, Appendix~\ref{appendix:case study} presents detailed case studies illustrating how length compression affects the quality and structure of generated responses in practice.

\subsection{Ablation Study on Reward Components}
\label{appendix: ablation on reward}

To understand the contribution of each reward component, we ablate SelfBudgeter by progressively adding format, correctness, PreB, and budget penalty terms. As shown in Table ~\ref{tab:ablation reward}, learning format only fails to preserve problem-solving ability: accuracy on all datasets collapses (e.g., GSM8K 1.64\%), despite producing shorter outputs. Adding correctness substantially restores accuracy, but compression becomes inconsistent, especially on harder datasets such as MATH500 and AIME2025, where reasoning lengths remain long (e.g., 5327 tokens on MATH500). Incorporating PreB improves budget adherence but harms performance on easier tasks, leading to notable accuracy drops on GSM8K and AIME2025. Adding budget penalty instead yields good compression on easy tasks but sacrifices performance on difficult ones (e.g., lower AIME2025 accuracy with long residual lengths). In contrast, SelfBudgeter, which integrates all reward components, achieves the best overall trade-off: high accuracy across datasets and strong, stable compression (e.g., 1231 tokens on GSM8K and 2327 tokens on MATH500). This demonstrates that the reward components are complementary—format ensures structure, correctness preserves reasoning quality, PreB stabilizes predicted budgets, and the budget penalty enforces compliance—and only their combined use yields robust reasoning-length control without degrading accuracy.

\subsection{$\alpha$ Scheduling}
\label{appendix: alpha scheduling}

We evaluated three $\alpha$-scheduling strategies: \textit{Fixed}, \textit{Cosine}, and \textit{Linear}. In the Fixed setting, $\alpha$ is held constant at $0.2$. For Cosine and Linear scheduling, $\alpha$ decreases from $6.0$ to $0.1$ over the training steps, following
\begin{equation}
\alpha_{\text{now}} = \alpha_{\text{start}} - \left(\alpha_{\text{start}} - \alpha_{\text{end}}\right) \cdot \frac{\text{step}_{\text{now}}}{\text{Total steps}}.
\end{equation}
\begin{equation}
\begin{aligned}
\alpha_{\text{cosine}} = \alpha_{\text{end}} + \dfrac{1}{2}(\alpha_{\text{start}} - \alpha_{\text{end}}) \times \\ 
\left(\cos\!\left(\pi \cdot \frac{\text{step}_{\text{now}}}{\text{Total steps}}\right) + 1\right).
\end{aligned}
\end{equation}
As shown in Table~\ref{tab:alpha-scheduling}, the Fixed strategy achieves the strongest length compression, with Cosine scheduling slightly weaker. However, both approaches tend to over-compress the reasoning trace, reducing accuracy on more difficult problems. In contrast, Linear scheduling maintains accuracy while still achieving a substantial compression ratio (over $61\%$), providing a more balanced solution that mitigates overthinking without degrading performance.

\begin{table*}[htbp]
\centering
\begin{tabular}{lcccc}
\toprule
\multirow{2}{*}{Model} & \multicolumn{2}{c}{GPQA} & \multicolumn{2}{c}{SCoRE} \\
\cmidrule(lr){2-3} \cmidrule(lr){4-5}
& Acc$\uparrow$ & Len$\downarrow$ & Acc$\uparrow$ & Len$\downarrow$ \\
\midrule
DeepSeek-R1-Distill-Qwen-1.5B & 33.04  & 11780.87 & 10.14 & 11695.94 \\
E1-Math-1.5B(0.5K,1K) & 26.34  & 1278.19 & 6.69 & 1272.04 \\
E1-Math-1.5B(4K,1K) & \textbf{34.30} & 2758.77 & 12.23 & 3327.37 \\
L1-Max(3600) & 31.92 & 3892.47 & 13.69 & 5145.91 \\
SelfBudgeter-1.5B & 30.65 & 3326.83 & \textbf{16.26} & 4491.35 \\
\bottomrule
\end{tabular}
\caption{Model performance comparison on GPQA and SCoRE test sets.}
\label{tab:gpqa_knowlogic_performance}
\end{table*}

\subsection{Generalization Ability}

To further validate the generalization ability of our method, we conduct additional experiments on two general-domain knowledge reasoning benchmarks, GPQA~\citep{rein2024gpqa} and SCoRE~\citep{zhan2025score}. As shown in Table \ref{tab:gpqa_knowlogic_performance}, the results exhibit a similar pattern to that observed on the math datasets. Compared with the long-reasoning baseline DeepSeek-R1-Distill-Qwen-1.5B, SelfBudgeter-1.5B substantially reduces reasoning length on both benchmarks while maintaining competitive accuracy. Although its GPQA accuracy is slightly below the strongest baseline, it remains comparable despite using far fewer reasoning tokens. On SCoRE, SelfBudgeter achieves the best accuracy among the compared methods, while also generating much shorter responses. These findings suggest that the advantages of SelfBudgeter are not limited to mathematical reasoning, but can also transfer to broader knowledge-intensive tasks, yielding a better trade-off between accuracy and inference efficiency.

\section{Conclusion}
\label{conclusion}
We propose the SelfBudgeter framework, which autonomously predicts required token budgets for reasoning while effectively adhering to self-imposed constraints, successfully optimizing the accuracy-response length trade-off. By leveraging SelfBudgeter's token budget predictions, users can anticipate total inference duration in advance, significantly enhancing user experience. In resource-efficient reasoning, SelfBudgeter demonstrates performance comparable to several existing methods, highlighting its potential for deployment in resource-constrained environments. Additionally, output length can be dynamically regulated through transformation functions when required. SelfBudgeter paves a promising pathway toward more efficient, controllable, and user-friendly reasoning models.

\section*{Limitations}


While this study presents promising results, it is not without certain inherent limitations that warrant consideration. First and foremost, the proposed methodology, though theoretically sound, may face practical challenges when applied to more complex or diverse real-world scenarios, as is often the case with cutting-edge research. Additionally, while extensive efforts have been made to ensure the robustness of our findings, certain external factors beyond the scope of this work could potentially influence the generalizability of the results. These limitations, however, do not detract from the overall contributions of this work but rather highlight opportunities for future research directions.

\section*{Ethical Considerations}
This work focuses on improving inference efficiency for large language models through budget-aware reasoning. It does not involve new data collection, human participants, or other direct ethical risks. We do not identify any specific ethical concerns beyond the general risks already associated with LLM deployment.

\section*{Acknowledgments}
This paper is supported by NSFC project 62476009 and the Open Project Fund of the State Key Laboratory of Multimedia Information Processing (Project No. SKLMIP-KF-2025-01).


\bibliography{custom}

@inproceedings{Wei2022ChainOT,
  author       = {Jason Wei and
                  Xuezhi Wang and
                  Dale Schuurmans and
                  Maarten Bosma and
                  Brian Ichter and
                  Fei Xia and
                  Ed H. Chi and
                  Quoc V. Le and
                  Denny Zhou},
  editor       = {Sanmi Koyejo and
                  S. Mohamed and
                  A. Agarwal and
                  Danielle Belgrave and
                  K. Cho and
                  A. Oh},
  title        = {Chain-of-Thought Prompting Elicits Reasoning in Large Language Models},
  booktitle    = {Advances in Neural Information Processing Systems 35: Annual Conference
                  on Neural Information Processing Systems 2022, NeurIPS 2022, New Orleans,
                  LA, USA, November 28 - December 9, 2022},
  year         = {2022},
  url          = {http://papers.nips.cc/paper\_files/paper/2022/hash/9d5609613524ecf4f15af0f7b31abca4-Abstract-Conference.html},
  bibsource    = {dblp computer science bibliography, https://dblp.org}
}

@misc{deepscaler2025,
  title={DeepScaleR: Surpassing O1-Preview with a 1.5B Model by Scaling RL},
  author={Michael Luo and Sijun Tan and Justin Wong and Xiaoxiang Shi and William Y. Tang and Manan Roongta and Colin Cai and Jeffrey Luo and Li Erran Li and Raluca Ada Popa and Ion Stoica},
  year={2025},
  note={Notion Blog}
}

@misc{openaiO1,
    title = {Learning to reason with LLMs},
    url = {https://openai.com/index/learning-to-reason-with-llms/},
    author = {OpenAI},
    month = {September},
    year = {2024}
}

@misc{qwq-32b-preview,
    title = {QwQ: Reflect Deeply on the Boundaries of the Unknown},
    url = {https://qwenlm.github.io/blog/qwq-32b-preview/},
    author = {Qwen},
    month = {November},
    year = {2024}
}

@misc{deepseekr1,
      title={DeepSeek-R1: Incentivizing Reasoning Capability in LLMs via Reinforcement Learning}, 
      author={DeepSeek-AI and Daya Guo and Dejian Yang and Haowei Zhang and Junxiao Song and Ruoyu Zhang and Runxin Xu and Qihao Zhu and Shirong Ma and Peiyi Wang and Xiao Bi and Xiaokang Zhang and Xingkai Yu and Yu Wu and Z. F. Wu and Zhibin Gou and Zhihong Shao and Zhuoshu Li and Ziyi Gao and Aixin Liu and Bing Xue and Bingxuan Wang and Bochao Wu and Bei Feng and Chengda Lu and Chenggang Zhao and Chengqi Deng and Chenyu Zhang and Chong Ruan and Damai Dai and Deli Chen and Dongjie Ji and Erhang Li and Fangyun Lin and Fucong Dai and Fuli Luo and Guangbo Hao and Guanting Chen and Guowei Li and H. Zhang and Han Bao and Hanwei Xu and Haocheng Wang and Honghui Ding and Huajian Xin and Huazuo Gao and Hui Qu and Hui Li and Jianzhong Guo and Jiashi Li and Jiawei Wang and Jingchang Chen and Jingyang Yuan and Junjie Qiu and Junlong Li and J. L. Cai and Jiaqi Ni and Jian Liang and Jin Chen and Kai Dong and Kai Hu and Kaige Gao and Kang Guan and Kexin Huang and Kuai Yu and Lean Wang and Lecong Zhang and Liang Zhao and Litong Wang and Liyue Zhang and Lei Xu and Leyi Xia and Mingchuan Zhang and Minghua Zhang and Minghui Tang and Meng Li and Miaojun Wang and Mingming Li and Ning Tian and Panpan Huang and Peng Zhang and Qiancheng Wang and Qinyu Chen and Qiushi Du and Ruiqi Ge and Ruisong Zhang and Ruizhe Pan and Runji Wang and R. J. Chen and R. L. Jin and Ruyi Chen and Shanghao Lu and Shangyan Zhou and Shanhuang Chen and Shengfeng Ye and Shiyu Wang and Shuiping Yu and Shunfeng Zhou and Shuting Pan and S. S. Li and Shuang Zhou and Shaoqing Wu and Shengfeng Ye and Tao Yun and Tian Pei and Tianyu Sun and T. Wang and Wangding Zeng and Wanjia Zhao and Wen Liu and Wenfeng Liang and Wenjun Gao and Wenqin Yu and Wentao Zhang and W. L. Xiao and Wei An and Xiaodong Liu and Xiaohan Wang and Xiaokang Chen and Xiaotao Nie and Xin Cheng and Xin Liu and Xin Xie and Xingchao Liu and Xinyu Yang and Xinyuan Li and Xuecheng Su and Xuheng Lin and X. Q. Li and Xiangyue Jin and Xiaojin Shen and Xiaosha Chen and Xiaowen Sun and Xiaoxiang Wang and Xinnan Song and Xinyi Zhou and Xianzu Wang and Xinxia Shan and Y. K. Li and Y. Q. Wang and Y. X. Wei and Yang Zhang and Yanhong Xu and Yao Li and Yao Zhao and Yaofeng Sun and Yaohui Wang and Yi Yu and Yichao Zhang and Yifan Shi and Yiliang Xiong and Ying He and Yishi Piao and Yisong Wang and Yixuan Tan and Yiyang Ma and Yiyuan Liu and Yongqiang Guo and Yuan Ou and Yuduan Wang and Yue Gong and Yuheng Zou and Yujia He and Yunfan Xiong and Yuxiang Luo and Yuxiang You and Yuxuan Liu and Yuyang Zhou and Y. X. Zhu and Yanhong Xu and Yanping Huang and Yaohui Li and Yi Zheng and Yuchen Zhu and Yunxian Ma and Ying Tang and Yukun Zha and Yuting Yan and Z. Z. Ren and Zehui Ren and Zhangli Sha and Zhe Fu and Zhean Xu and Zhenda Xie and Zhengyan Zhang and Zhewen Hao and Zhicheng Ma and Zhigang Yan and Zhiyu Wu and Zihui Gu and Zijia Zhu and Zijun Liu and Zilin Li and Ziwei Xie and Ziyang Song and Zizheng Pan and Zhen Huang and Zhipeng Xu and Zhongyu Zhang and Zhen Zhang},
      year={2025},
      eprint={2501.12948},
      archivePrefix={arXiv},
      primaryClass={cs.CL},
      url={https://arxiv.org/abs/2501.12948}, 
}

@article{test-time-compute,
  author       = {Charlie Snell and
                  Jaehoon Lee and
                  Kelvin Xu and
                  Aviral Kumar},
  title        = {Scaling {LLM} Test-Time Compute Optimally can be More Effective than
                  Scaling Model Parameters},
  journal      = {CoRR},
  volume       = {abs/2408.03314},
  year         = {2024},
  url          = {https://doi.org/10.48550/arXiv.2408.03314},
  doi          = {10.48550/ARXIV.2408.03314},
  eprinttype    = {arXiv},
  eprint       = {2408.03314},
  bibsource    = {dblp computer science bibliography, https://dblp.org}
}

@misc{chen2024think23overthinkingo1like,
      title={Do NOT Think That Much for 2+3=? On the Overthinking of o1-Like LLMs}, 
      author={Xingyu Chen and Jiahao Xu and Tian Liang and Zhiwei He and Jianhui Pang and Dian Yu and Linfeng Song and Qiuzhi Liu and Mengfei Zhou and Zhuosheng Zhang and Rui Wang and Zhaopeng Tu and Haitao Mi and Dong Yu},
      year={2024},
      eprint={2412.21187},
      archivePrefix={arXiv},
      primaryClass={cs.CL},
      url={https://arxiv.org/abs/2412.21187}, 
}

@misc{qwen2025qwen25technicalreport,
      title={Qwen2.5 Technical Report}, 
      author={Qwen and An Yang and Baosong Yang and Beichen Zhang and Binyuan Hui and Bo Zheng and Bowen Yu and Chengyuan Li and Dayiheng Liu and Fei Huang and Haoran Wei and Huan Lin and Jian Yang and Jianhong Tu and Jianwei Zhang and Jianxin Yang and Jiaxi Yang and Jingren Zhou and Junyang Lin and Kai Dang and Keming Lu and Keqin Bao and Kexin Yang and Le Yu and Mei Li and Mingfeng Xue and Pei Zhang and Qin Zhu and Rui Men and Runji Lin and Tianhao Li and Tianyi Tang and Tingyu Xia and Xingzhang Ren and Xuancheng Ren and Yang Fan and Yang Su and Yichang Zhang and Yu Wan and Yuqiong Liu and Zeyu Cui and Zhenru Zhang and Zihan Qiu},
      year={2025},
      eprint={2412.15115},
      archivePrefix={arXiv},
      primaryClass={cs.CL},
      url={https://arxiv.org/abs/2412.15115}, 
}

@article{conciseThoughts,
  author       = {Sania Nayab and
                  Giulio Rossolini and
                  Giorgio C. Buttazzo and
                  Nicolamaria Manes and
                  Fabrizio Giacomelli},
  title        = {Concise Thoughts: Impact of Output Length on {LLM} Reasoning and Cost},
  journal      = {CoRR},
  volume       = {abs/2407.19825},
  year         = {2024},
  url          = {https://doi.org/10.48550/arXiv.2407.19825},
  doi          = {10.48550/ARXIV.2407.19825},
  eprinttype    = {arXiv},
  eprint       = {2407.19825},
  bibsource    = {dblp computer science bibliography, https://dblp.org}
}

@article{DBLP:whenmoreisless,
  author       = {Yuyang Wu and
                  Yifei Wang and
                  Tianqi Du and
                  Stefanie Jegelka and
                  Yisen Wang},
  title        = {When More is Less: Understanding Chain-of-Thought Length in LLMs},
  journal      = {CoRR},
  volume       = {abs/2502.07266},
  year         = {2025},
  url          = {https://doi.org/10.48550/arXiv.2502.07266},
  doi          = {10.48550/ARXIV.2502.07266},
  eprinttype    = {arXiv},
  eprint       = {2502.07266},
  bibsource    = {dblp computer science bibliography, https://dblp.org}
}

@article{sui2025stop,
  title={Stop Overthinking: A Survey on Efficient Reasoning for Large Language Models},
  author={Sui, Yang and Chuang, Yu-Neng and Wang, Guanchu and Zhang, Jiamu and Zhang, Tianyi and Yuan, Jiayi and Liu, Hongyi and Wen, Andrew and Chen, Hanjie and Hu, Xia and others},
  journal={arXiv preprint arXiv:2503.16419},
  year={2025}
}

@article{han2024token,
  title={Token-budget-aware llm reasoning},
  author={Han, Tingxu and Wang, Zhenting and Fang, Chunrong and Zhao, Shiyu and Ma, Shiqing and Chen, Zhenyu},
  journal={arXiv preprint arXiv:2412.18547},
  year={2024}
}

@article{DBLP:tokencomplexity,
  author       = {Ayeong Lee and
                  Ethan Che and
                  Tianyi Peng},
  title        = {How Well do LLMs Compress Their Own Chain-of-Thought? {A} Token Complexity
                  Approach},
  journal      = {CoRR},
  volume       = {abs/2503.01141},
  year         = {2025},
  url          = {https://doi.org/10.48550/arXiv.2503.01141},
  doi          = {10.48550/ARXIV.2503.01141},
  eprinttype    = {arXiv},
  eprint       = {2503.01141},
  bibsource    = {dblp computer science bibliography, https://dblp.org}
}

@inproceedings{DBLP:ccot,
  author       = {Matthew Renze and
                  Erhan Guven},
  title        = {The Benefits of a Concise Chain of Thought on Problem-Solving in Large
                  Language Models},
  booktitle    = {2nd International Conference on Foundation and Large Language Models,
                  {FLLM} 2024, Dubai, United Arab Emirates, November 26-29, 2024},
  pages        = {476--483},
  publisher    = {{IEEE}},
  year         = {2024},
  url          = {https://doi.org/10.1109/FLLM63129.2024.10852493},
  doi          = {10.1109/FLLM63129.2024.10852493},
  bibsource    = {dblp computer science bibliography, https://dblp.org}
}

@article{DBLP:cod,
  author       = {Silei Xu and
                  Wenhao Xie and
                  Lingxiao Zhao and
                  Pengcheng He},
  title        = {Chain of Draft: Thinking Faster by Writing Less},
  journal      = {CoRR},
  volume       = {abs/2502.18600},
  year         = {2025},
  url          = {https://doi.org/10.48550/arXiv.2502.18600},
  doi          = {10.48550/ARXIV.2502.18600},
  eprinttype    = {arXiv},
  eprint       = {2502.18600},
  bibsource    = {dblp computer science bibliography, https://dblp.org}
}

@article{DBLP:self-training,
  author       = {Tergel Munkhbat and
                  Namgyu Ho and
                  Seo Hyun Kim and
                  Yongjin Yang and
                  Yujin Kim and
                  Se{-}Young Yun},
  title        = {Self-Training Elicits Concise Reasoning in Large Language Models},
  journal      = {CoRR},
  volume       = {abs/2502.20122},
  year         = {2025},
  url          = {https://doi.org/10.48550/arXiv.2502.20122},
  doi          = {10.48550/ARXIV.2502.20122},
  eprinttype    = {arXiv},
  eprint       = {2502.20122},
  bibsource    = {dblp computer science bibliography, https://dblp.org}
}

@article{DBLP:cot-valve,
  author       = {Xinyin Ma and
                  Guangnian Wan and
                  Runpeng Yu and
                  Gongfan Fang and
                  Xinchao Wang},
  title        = {CoT-Valve: Length-Compressible Chain-of-Thought Tuning},
  journal      = {CoRR},
  volume       = {abs/2502.09601},
  year         = {2025},
  url          = {https://doi.org/10.48550/arXiv.2502.09601},
  doi          = {10.48550/ARXIV.2502.09601},
  eprinttype    = {arXiv},
  eprint       = {2502.09601},
  bibsource    = {dblp computer science bibliography, https://dblp.org}
}

@inproceedings{DBLP:skipsteps,
  author       = {Tengxiao Liu and
                  Qipeng Guo and
                  Xiangkun Hu and
                  Cheng Jiayang and
                  Yue Zhang and
                  Xipeng Qiu and
                  Zheng Zhang},
  editor       = {Amir Globersons and
                  Lester Mackey and
                  Danielle Belgrave and
                  Angela Fan and
                  Ulrich Paquet and
                  Jakub M. Tomczak and
                  Cheng Zhang},
  title        = {Can Language Models Learn to Skip Steps?},
  booktitle    = {Advances in Neural Information Processing Systems 38: Annual Conference
                  on Neural Information Processing Systems 2024, NeurIPS 2024, Vancouver,
                  BC, Canada, December 10 - 15, 2024},
  year         = {2024},
  url          = {http://papers.nips.cc/paper\_files/paper/2024/hash/504fa7e518da9d1b53a233ed20a38b46-Abstract-Conference.html},
  bibsource    = {dblp computer science bibliography, https://dblp.org}
}

@article{DBLP:c3ot,
  author       = {Yu Kang and
                  Xianghui Sun and
                  Liangyu Chen and
                  Wei Zou},
  title        = {C3oT: Generating Shorter Chain-of-Thought without Compromising Effectiveness},
  journal      = {CoRR},
  volume       = {abs/2412.11664},
  year         = {2024},
  url          = {https://doi.org/10.48550/arXiv.2412.11664},
  doi          = {10.48550/ARXIV.2412.11664},
  eprinttype    = {arXiv},
  eprint       = {2412.11664},
  bibsource    = {dblp computer science bibliography, https://dblp.org}
}

@article{DBLP:tokenskip,
  author       = {Heming Xia and
                  Yongqi Li and
                  Chak Tou Leong and
                  Wenjie Wang and
                  Wenjie Li},
  title        = {TokenSkip: Controllable Chain-of-Thought Compression in LLMs},
  journal      = {CoRR},
  volume       = {abs/2502.12067},
  year         = {2025},
  url          = {https://doi.org/10.48550/arXiv.2502.12067},
  doi          = {10.48550/ARXIV.2502.12067},
  eprinttype    = {arXiv},
  eprint       = {2502.12067},
  bibsource    = {dblp computer science bibliography, https://dblp.org}
}

@article{aggarwal2025l1,
  title={L1: Controlling how long a reasoning model thinks with reinforcement learning},
  author={Aggarwal, Pranjal and Welleck, Sean},
  journal={arXiv preprint arXiv:2503.04697},
  year={2025}
}

@article{DBLP:trainingllmtoreasonefficiently,
  author       = {Daman Arora and
                  Andrea Zanette},
  title        = {Training Language Models to Reason Efficiently},
  journal      = {CoRR},
  volume       = {abs/2502.04463},
  year         = {2025},
  url          = {https://doi.org/10.48550/arXiv.2502.04463},
  doi          = {10.48550/ARXIV.2502.04463},
  eprinttype    = {arXiv},
  eprint       = {2502.04463},
  bibsource    = {dblp computer science bibliography, https://dblp.org}
}

@article{DBLP:o1-pruner,
  author       = {Haotian Luo and
                  Li Shen and
                  Haiying He and
                  Yibo Wang and
                  Shiwei Liu and
                  Wei Li and
                  Naiqiang Tan and
                  Xiaochun Cao and
                  Dacheng Tao},
  title        = {O1-Pruner: Length-Harmonizing Fine-Tuning for O1-Like Reasoning Pruning},
  journal      = {CoRR},
  volume       = {abs/2501.12570},
  year         = {2025},
  url          = {https://doi.org/10.48550/arXiv.2501.12570},
  doi          = {10.48550/ARXIV.2501.12570},
  eprinttype    = {arXiv},
  eprint       = {2501.12570},
  bibsource    = {dblp computer science bibliography, https://dblp.org}
}

@article{chen2025towards,
  title={Towards reasoning era: A survey of long chain-of-thought for reasoning large language models},
  author={Chen, Qiguang and Qin, Libo and Liu, Jinhao and Peng, Dengyun and Guan, Jiannan and Wang, Peng and Hu, Mengkang and Zhou, Yuhang and Gao, Te and Che, Wangxiang},
  journal={arXiv preprint arXiv:2503.09567},
  year={2025}
}

@article{DBLP:demystifying,
  author       = {Edward Y. Chang and
                  Yuxuan Tong and
                  Morry Niu and
                  Graham Neubig and
                  Xiang Yue},
  title        = {Demystifying Long Chain-of-Thought Reasoning in LLMs},
  journal      = {CoRR},
  volume       = {abs/2502.03373},
  year         = {2025},
  url          = {https://doi.org/10.48550/arXiv.2502.03373},
  doi          = {10.48550/ARXIV.2502.03373},
  eprinttype    = {arXiv},
  eprint       = {2502.03373},
  bibsource    = {dblp computer science bibliography, https://dblp.org}
}

@article{aytes2025sketch,
  title={Sketch-of-thought: Efficient llm reasoning with adaptive cognitive-inspired sketching},
  author={Aytes, Simon A and Baek, Jinheon and Hwang, Sung Ju},
  journal={arXiv preprint arXiv:2503.05179},
  year={2025}
}

@article{chuang2025confident,
  title={Confident or Seek Stronger: Exploring Uncertainty-Based On-device LLM Routing From Benchmarking to Generalization},
  author={Chuang, Yu-Neng and Yu, Leisheng and Wang, Guanchu and Zhang, Lizhe and Liu, Zirui and Cai, Xuanting and Sui, Yang and Braverman, Vladimir and Hu, Xia},
  journal={arXiv preprint arXiv:2502.04428},
  year={2025}
}

@misc{eurus7b,
      title={Process Reinforcement through Implicit Rewards}, 
      author={Ganqu Cui and Lifan Yuan and Zefan Wang and Hanbin Wang and Wendi Li and Bingxiang He and Yuchen Fan and Tianyu Yu and Qixin Xu and Weize Chen and Jiarui Yuan and Huayu Chen and Kaiyan Zhang and Xingtai Lv and Shuo Wang and Yuan Yao and Xu Han and Hao Peng and Yu Cheng and Zhiyuan Liu and Maosong Sun and Bowen Zhou and Ning Ding},
      year={2025},
      eprint={2502.01456},
      archivePrefix={arXiv},
      primaryClass={cs.LG},
      url={https://arxiv.org/abs/2502.01456}, 
}

@misc{simplerl7b,
      title={DeepSeekMath: Pushing the Limits of Mathematical Reasoning in Open Language Models}, 
      author={Zhihong Shao and Peiyi Wang and Qihao Zhu and Runxin Xu and Junxiao Song and Xiao Bi and Haowei Zhang and Mingchuan Zhang and Y. K. Li and Y. Wu and Daya Guo},
      year={2024},
      eprint={2402.03300},
      archivePrefix={arXiv},
      primaryClass={cs.CL},
      url={https://arxiv.org/abs/2402.03300}, 
}

@article{DBLP:gsm,
  author       = {Karl Cobbe and
                  Vineet Kosaraju and
                  Mohammad Bavarian and
                  Mark Chen and
                  Heewoo Jun and
                  Lukasz Kaiser and
                  Matthias Plappert and
                  Jerry Tworek and
                  Jacob Hilton and
                  Reiichiro Nakano and
                  Christopher Hesse and
                  John Schulman},
  title        = {Training Verifiers to Solve Math Word Problems},
  journal      = {CoRR},
  volume       = {abs/2110.14168},
  year         = {2021},
  url          = {https://arxiv.org/abs/2110.14168},
  eprinttype    = {arXiv},
  eprint       = {2110.14168},
  bibsource    = {dblp computer science bibliography, https://dblp.org}
}

@article{DBLP:self-ref,
  author       = {Yu{-}Neng Chuang and
                  Helen Zhou and
                  Prathusha Kameswara Sarma and
                  Parikshit Gopalan and
                  John Boccio and
                  Sara Bolouki and
                  Xia Hu},
  title        = {Learning to Route with Confidence Tokens},
  journal      = {CoRR},
  volume       = {abs/2410.13284},
  year         = {2024},
  url          = {https://doi.org/10.48550/arXiv.2410.13284},
  doi          = {10.48550/ARXIV.2410.13284},
  eprinttype    = {arXiv},
  eprint       = {2410.13284},
  bibsource    = {dblp computer science bibliography, https://dblp.org}
}

@misc{xu2025scalablechainthoughtselastic,
      title={Scalable Chain of Thoughts via Elastic Reasoning}, 
      author={Yuhui Xu and Hanze Dong and Lei Wang and Doyen Sahoo and Junnan Li and Caiming Xiong},
      year={2025},
      eprint={2505.05315},
      archivePrefix={arXiv},
      primaryClass={cs.LG},
      url={https://arxiv.org/abs/2505.05315}, 
}

@article{DBLP:routellm,
  author       = {Isaac Ong and
                  Amjad Almahairi and
                  Vincent Wu and
                  Wei{-}Lin Chiang and
                  Tianhao Wu and
                  Joseph E. Gonzalez and
                  M. Waleed Kadous and
                  Ion Stoica},
  title        = {RouteLLM: Learning to Route LLMs with Preference Data},
  journal      = {CoRR},
  volume       = {abs/2406.18665},
  year         = {2024},
  url          = {https://doi.org/10.48550/arXiv.2406.18665},
  doi          = {10.48550/ARXIV.2406.18665},
  eprinttype    = {arXiv},
  eprint       = {2406.18665},
  bibsource    = {dblp computer science bibliography, https://dblp.org}
}

@article{yang2025towards,
  title={Towards thinking-optimal scaling of test-time compute for llm reasoning},
  author={Yang, Wenkai and Ma, Shuming and Lin, Yankai and Wei, Furu},
  journal={arXiv preprint arXiv:2502.18080},
  year={2025}
}

@misc{pan2025specreasonfastaccurateinferencetime,
      title={SpecReason: Fast and Accurate Inference-Time Compute via Speculative Reasoning}, 
      author={Rui Pan and Yinwei Dai and Zhihao Zhang and Gabriele Oliaro and Zhihao Jia and Ravi Netravali},
      year={2025},
      eprint={2504.07891},
      archivePrefix={arXiv},
      primaryClass={cs.LG},
      url={https://arxiv.org/abs/2504.07891}, 
}

@article{DBLP:s1dataset,
  author       = {Niklas Muennighoff and
                  Zitong Yang and
                  Weijia Shi and
                  Xiang Lisa Li and
                  Li Fei{-}Fei and
                  Hannaneh Hajishirzi and
                  Luke Zettlemoyer and
                  Percy Liang and
                  Emmanuel J. Cand{\`{e}}s and
                  Tatsunori Hashimoto},
  title        = {s1: Simple test-time scaling},
  journal      = {CoRR},
  volume       = {abs/2501.19393},
  year         = {2025},
  url          = {https://doi.org/10.48550/arXiv.2501.19393},
  doi          = {10.48550/ARXIV.2501.19393},
  eprinttype    = {arXiv},
  eprint       = {2501.19393},
  bibsource    = {dblp computer science bibliography, https://dblp.org}
}

@article{Still3,
  title={An Empirical Study on Eliciting and Improving R1-like Reasoning Models},
  author={Chen, Zhipeng and Min, Yingqian and Zhang, Beichen  and Chen, Jie and Jiang, Jinhao and Cheng, Daixuan and Zhao, Wayne Xin and Liu, Zheng and Miao, Xu and Lu, Yang and Fang, Lei and Wang, Zhongyuan and Wen, Ji-Rong},
  journal={arXiv preprint arXiv:2503.04548},
  year={2025}
}

@misc{yang2025dynamicearlyexitreasoning,
      title={Dynamic Early Exit in Reasoning Models}, 
      author={Chenxu Yang and Qingyi Si and Yongjie Duan and Zheliang Zhu and Chenyu Zhu and Qiaowei Li and Zheng Lin and Li Cao and Weiping Wang},
      year={2025},
      eprint={2504.15895},
      archivePrefix={arXiv},
      primaryClass={cs.CL},
      url={https://arxiv.org/abs/2504.15895}, 
}

@inproceedings{DBLP:math,
  author       = {Dan Hendrycks and
                  Collin Burns and
                  Saurav Kadavath and
                  Akul Arora and
                  Steven Basart and
                  Eric Tang and
                  Dawn Song and
                  Jacob Steinhardt},
  editor       = {Joaquin Vanschoren and
                  Sai{-}Kit Yeung},
  title        = {Measuring Mathematical Problem Solving With the {MATH} Dataset},
  booktitle    = {Proceedings of the Neural Information Processing Systems Track on
                  Datasets and Benchmarks 1, NeurIPS Datasets and Benchmarks 2021, December
                  2021, virtual},
  year         = {2021},
  url          = {https://datasets-benchmarks-proceedings.neurips.cc/paper/2021/hash/be83ab3ecd0db773eb2dc1b0a17836a1-Abstract-round2.html},
  bibsource    = {dblp computer science bibliography, https://dblp.org}
}

@misc{numina_math_datasets,
  author = {Jia Li and Edward Beeching and Lewis Tunstall and Ben Lipkin and Roman Soletskyi and Shengyi Costa Huang and Kashif Rasul and Longhui Yu and Albert Jiang and Ziju Shen and Zihan Qin and Bin Dong and Li Zhou and Yann Fleureau and Guillaume Lample and Stanislas Polu},
  title = {NuminaMath},
  year = {2024},
  publisher = {Numina},
  journal = {Hugging Face repository},
  url = {https://huggingface.co/datasets/AI-MO/NuminaMath-CoT}
}

@dataset{aime_1983_2024,
  author = {Hemish Veeraboina},
  title = {AIME Problem Set 1983-2024},
  year = {2023},
  publisher = {Kaggle},
  url = {https://www.kaggle.com/datasets/hemishveeraboina/aime-problem-set-1983-2024}
}

@inproceedings{zeng-etal-2025-revisiting,
    title = "Revisiting the Test-Time Scaling of o1-like Models: Do they Truly Possess Test-Time Scaling Capabilities?",
    author = "Zeng, Zhiyuan  and
      Cheng, Qinyuan  and
      Yin, Zhangyue  and
      Zhou, Yunhua  and
      Qiu, Xipeng",
    editor = "Che, Wanxiang  and
      Nabende, Joyce  and
      Shutova, Ekaterina  and
      Pilehvar, Mohammad Taher",
    booktitle = "Proceedings of the 63rd Annual Meeting of the Association for Computational Linguistics (Volume 1: Long Papers)",
    month = jul,
    year = "2025",
    address = "Vienna, Austria",
    publisher = "Association for Computational Linguistics",
    url = "https://aclanthology.org/2025.acl-long.232/",
    doi = "10.18653/v1/2025.acl-long.232",
    pages = "4651--4665",
    ISBN = "979-8-89176-251-0"
}

@inproceedings{rein2024gpqa,
  title={Gpqa: A graduate-level google-proof q\&a benchmark},
  author={Rein, David and Hou, Betty Li and Stickland, Asa Cooper and Petty, Jackson and Pang, Richard Yuanzhe and Dirani, Julien and Michael, Julian and Bowman, Samuel R},
  booktitle={First conference on language modeling}
}

@article{zhan2025score,
  title={SCoRE: Benchmarking long-chain reasoning in commonsense scenarios},
  author={Zhan, Weidong and Wang, Yue and Hu, Nan and Xiao, Liming and Ma, Jingyuan and Qin, Yuhang and Li, Zheng and Yang, Yixin and Deng, Sirui and Ding, Jinkun and others},
  journal={arXiv preprint arXiv:2503.06218},
  year={2025}
}

\newpage

\appendix
\section{Training Details}
\label{appendix:training details}
\subsection{Experimental Environments}
Our server is equipped with two 80GB A100 GPUs and two 45GB A40 GPUs. We conducted fine-tuning experiments and inference tests on the two A40 GPUs, while the GRPO training was performed on the two A100 GPUs.

\subsection{Parameter Settings}

In the fine-tuning training during the cold-start phase, our parameter settings are configured as follows. The sequence length is capped at 16,384, with a per-device training and evaluation batch size of 1, while gradient accumulation (2 steps) is employed to alleviate GPU memory constraints. A cosine learning rate scheduler is adopted with a 10\% warm-up ratio and a base learning rate of 5e-5. The model is trained for 1 epoch, with 10\% of the training set allocated for validation. The model checkpoints are saved and evaluated every 500 steps, and the best-performing checkpoint is retained.

In the GRPO (Global Reward Policy Optimization) training, our parameter configuration is set as follows. The training and validation batch sizes are set to 128 and 1,250, respectively, with maximum prompt and response lengths of 1,024 and 32,000 tokens. The Actor model employs a learning rate of 1e-6, dynamic batching (up to 24K tokens per GPU), and a KL divergence loss (coefficient 0.001), with gradient checkpointing and FSDP (Fully Sharded Data Parallel) distributed training enabled (parameter offloading disabled). During the Rollout phase, the vLLM inference engine is utilized with tensor parallelism (TP=2) and 80\% GPU memory utilization, generating 5 responses per round. Global settings include 3 training epochs, a checkpoint-saving interval of 50 steps, and a KL control coefficient of 0.001, executed on a single node with dual GPUs. Key hyperparameters involved in the reward function are specified in Table \ref{tab:hps for reward func}.
\begin{table}[h]
\centering
\begin{tabular}{ccccc}
\toprule
\textbf{Params} & \textbf{$C=0$} & \textbf{$C=1$} & \textbf{Params} & \textbf{Value}\\
\midrule
$s_{\text{min}}$ & -0.5 & 0.5 & $r_f$ & -1\\
$s_{\text{max}}$ & 0 & 1 & $r_b$ & -0.4\\
\bottomrule
\end{tabular}
\caption{Hyperparameter Settings}
\label{tab:hps for reward func}
\end{table}

For the GSM-initialized SelfBudgeter, we select the checkpoint after 699 training steps when $\alpha$ was set to 0.2, and the checkpoint after 575 steps when $\alpha$ was 0.5. For the s1k-initialized SelfBudgeter, we choose the checkpoint after 475 training steps with $\alpha=0.2$, and the checkpoint after 500 steps with $\alpha=0.5$. For SelfBudgeter-1.5B, SelfBudgeter-7B, and L1-Max, we choose the checkpoint after 699 training steps.

\section{Cold-start Data Selection}
\label{appendix:cold-start}
\begin{figure}[h!]
    \centering
        \begin{tcolorbox}[colback=blue!5!white, colframe=blue!75!black, title=Prompt Template]
            Answer the given question. You should first estimate the total number of tokens you will need to answer this question based on its difficulty. Then you think about the reasoning process in the mind and provide the user with the answer. The token budget and whole solution are enclosed within <budget></budget> and <solution> </solution> tags, respectively, i.e., <budget> token budget here, just an integer </budget><solution> solution here, please output the final answer within \textbackslash boxed\{\} </solution>.
            
            Question: 
        \end{tcolorbox}
    \caption{The prompt template used in the cold-start stage.}
    \label{fig:prompt-template}
\end{figure}

\begin{table*}[htbp]
\centering
\begin{tabular}{lcccccc}
\toprule
\multirow{2}{*}{Model} & \multicolumn{3}{c}{GSM8K} & \multicolumn{3}{c}{MATH} \\
\cmidrule(lr){2-4} \cmidrule(lr){5-7}
 & Acc$\uparrow$ & Len$\downarrow$ & Mat$\uparrow$ & Acc$\uparrow$ & Len$\downarrow$ & Mat$\uparrow$ \\
\midrule
Cold Start (GSM) & 71.95 & 1003.79 & 85.82 & 64.74 & 3043.29 & 41.16\\
SelfBudgeter (GSM, $\alpha=0.2$) & 76.27 & 523.77 & \textbf{97.65} & 63.46 & 779.54 & 95.82\\
SelfBudgeter (GSM, $\alpha=0.5$) & 74.68 & 520.82 & 96.97 & 63.78 & 777.80 & \textbf{96.66}\\
Cold Start (s1k) & 82.49 & 1983.29 & 21.76 & 76.64 & 4001.29 & 23.28 \\
SelfBudgeter (s1k, $\alpha=0.2$) & \textbf{81.50} & 662.08 & 70.74 & \textbf{74.18} & 919.27 & 78.36\\
SelfBudgeter (s1k, $\alpha=0.5$) & 80.44 & 719.36 & 71.19 & 72.60 & 1022.99 & 79.76 \\
\bottomrule
\end{tabular}
\caption{Model performance comparison on GSM8K and MATH test sets, showing accuracy (Acc/\%), average response length (Len/tokens) and matching rate between token limits and response length (Mat/\%). The SelfBudgeter variants with different cold-start data and $\alpha$ parameters are contrasted with baseline models.}
\label{tab:cold start data performance}
\end{table*}
The choice of initialization data substantially impacts model performance. SelfBudgeters initialized with the \textbf{s1k} dataset outperform their GSM-initialized SelfBudgeters by 8.82--10.72 percentage points on MATH (74.18\% vs. 63.46\% for $\alpha=0.2$) and 5.23--5.76 percentage points on GSM8K (80.44\% vs. 74.68\% for $\alpha=0.5$). While SelfBudgeters with GSM-initialized exhibit lower accuracy, they generate significantly more concise responses compared to s1k-initialized SelfBudgeters. Specifically, GSM-initialized SelfBudgeters reduce response length by approximately 15--24\% on MATH and by 21--28\% on GSM8K. This performance gap highlights the importance of high-quality initialization for the budgeting mechanism.

As shown in Table \ref{tab:cold start data performance}, significant performance variations exist between models fine-tuned with different cold-start datasets. The s1k-fine-tuned model demonstrates superior accuracy over the GSM-fine-tuned counterpart, achieving 10.54\% and 11.90\% higher accuracy on GSM8K and MATH respectively. However, this comes at the cost of substantially longer responses, with the s1k model generating 97.58\% and 31.48\% lengthier outputs on GSM8K and MATH. This discrepancy stems from the s1k dataset's responses being generated by DeepSeek-R1, which produces higher-quality outputs than those self-generated by DeepSeek-R1-Distill-Qwen-1.5B. Additionally, the s1k dataset's average length of 7,677.43 tokens (we only retained correct responses under 16,000 tokens) vastly exceeds GSM8K's 837.14 tokens, explaining the dramatic difference in response lengths after fine-tuning. These factors substantially influence SelfBudgeter's final performance, as evidenced by: (1) SelfBudgeter's accuracy closely mirroring that of its fine-tuned base model, and (2) the response length relationships and matching rate relationships between different SelfBudgeter variants remaining consistent with their respective cold-start models.

\section{Prefilled Token Budget Following}
\label{appendix:prefill budget}

\begin{figure*}[htbp]
    \centering
    \includegraphics[width=1\linewidth]{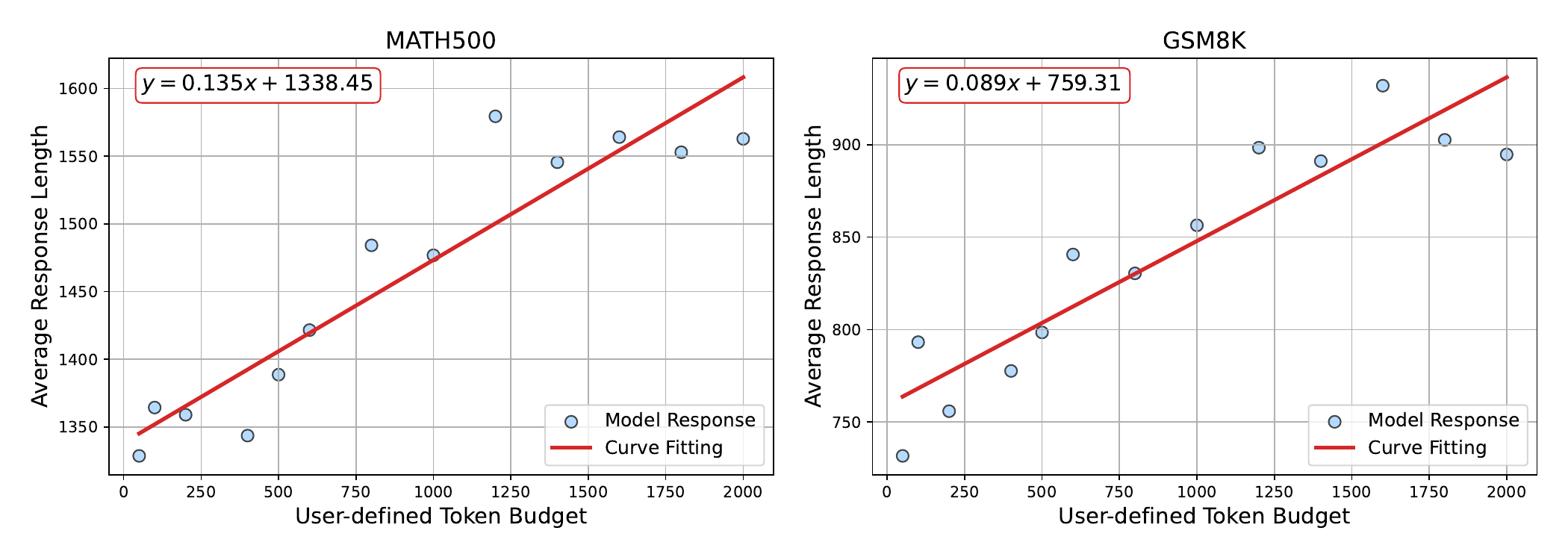} 
    \caption{Relationship between user-defined token budgets and SelfBudgeter average response lengths with curve fitting analysis.}
    \label{fig:budget prefill}
\end{figure*}

To systematically evaluate model performance under user-defined token budget constraints, we conduct quantitative analysis using SelfBudgeter with GSM initialization and hyperparameter $\alpha=0.2$ on both MATH500 dataset and GSM8K test set. In the experimental design, fixed token budgets were pre-filled in the <budget> field of training templates, with empirical results obtained by measuring average generated response lengths. We evaluated SelfBudgeter's performance with user-defined token budgets ranging from 50 to 2000 (specifically: 50, 100, 200, 400, 500, 600, 800, 1000, 1200, 1400, 1600, 1800, and 2000), as shown in Figure \ref{fig:budget prefill}.

Regression intercepts effectively reflect problem complexity, where GSM8K's simpler questions yield significantly smaller intercepts. Despite a moderate slope, SelfBudgeter demonstrates robust budget adaptability, maintaining a stable positive correlation between user-defined budgets and output lengths. This linear relationship enables deterministic length control through derived transformation functions.

\section{Adaptive Budget Allocation}

\label{appendix: adaptive budget allocation}
\begin{figure*}[htbp]
    \centering
    \includegraphics[width=0.8\linewidth]{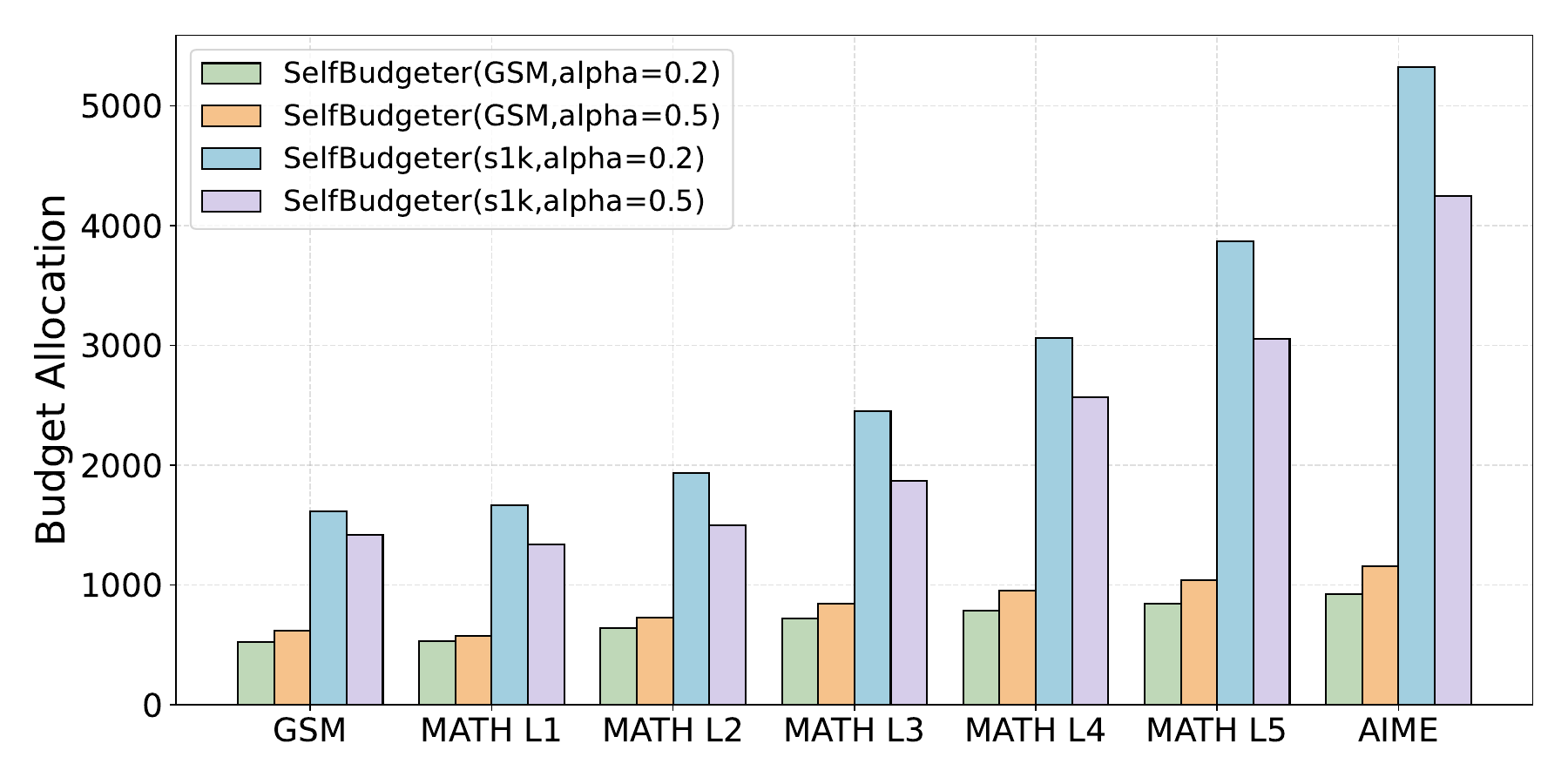}
    \caption{Token budget allocation patterns across problem difficulty levels for four SelfBudgeter-1.5B configurations (initialized on GSM8K/s1k with $\alpha$=0.2/0.5). All variants exhibit monotonic budget escalation with increasing task complexity (GSM8K, MATH Level 1-5, AIME2024), confirming robust cross-configuration alignment between computational investment and intrinsic problem difficulty.}
    \label{fig:budget allocation}
\end{figure*}

To investigate SelfBudgeter's capacity for difficulty-aware budget allocation, we conduct empirical evaluations across three mathematical reasoning benchmarks with inherent complexity gradients: GSM8K, MATH, and AIME 2024. Our experimental framework systematically evaluates four architectural variants combining cold-start initialization strategies (GSM8K vs. s1k) with $\alpha$ hyperparameter values (0.2 vs. 0.5).


Figure~\ref{fig:budget allocation} shows a consistent positive correlation between problem complexity and allocated token budgets across all model variants, demonstrating SelfBudgeter’s ability to scale computation with task difficulty. The near-linear allocation across difficulty tiers highlights its emergent capacity for intrinsic difficulty estimation, while the minimal variance across configurations indicates robust and generalized learning of task-complexity metrics rather than configuration-specific artifacts.

\section{Budget-Constrained Generation}
\label{appendix:budget constrained generation}
\begin{figure*}
    \centering
    \includegraphics[width=1\linewidth]{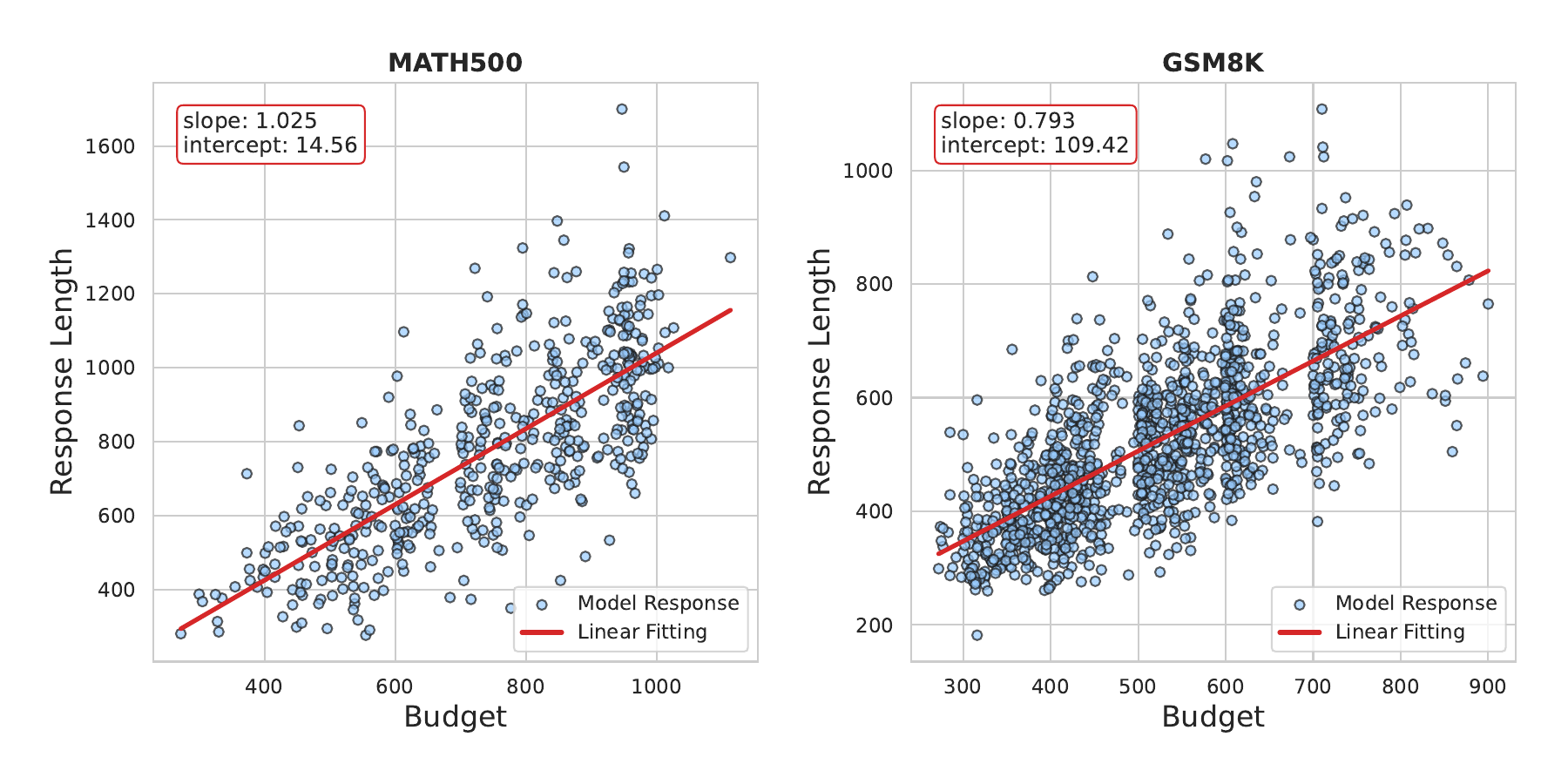} 
    \caption{Linear regression analysis of token budget prediction versus actual response length for SelfBudgeter on the MATH500 dataset (left) and GSM8K test set (right). The figure demonstrates SelfBudgeter with GSM initialization and hyperparameter $\alpha=0.2$.}
    \label{fig:budget length relation}
\end{figure*}
To systematically evaluate the generation capability of SelfBudgeter under budget constraints, this study employs linear regression modeling to quantitatively analyze the mapping relationship between predicted token budgets and actual response lengths. We conduct a quantitative analysis on the MATH500 dataset and GSM8K test set using linear regression to investigate the mapping between predicted budgets and actual response lengths (as shown in Figure \ref{fig:budget length relation}). On MATH500 dataset, the least squares fitting yields a slope of 1.025 (95\% CI [0.9466, 1.1042]). On GSM8K test set, the least squares fitting yields a slope of 0.793 (95\% CI [0.7512, 0.8354]). The slope coefficient approaching unity validates the efficacy of the budget control mechanism, indicating that each 1-token increase in the predicted budget corresponds to an average increase of about 1 token in output. 

Quantitative results demonstrate that 96\% of generated responses exhibit relative deviations $\leq 50\%$  from the target token budget, with 65.40\% achieving tighter deviations $\leq 20\%$. Extended experiments on full benchmark datasets reveal that 97.65\% (GSM8K) and 95.82\% (MATH) of samples satisfy the $\leq 50\%$  relative deviation constraint. Notably, the model’s budget adherence is influenced by the cold-start dataset and hyperparameter $\alpha$. The optimized SelfBudgeter configuration (initialized with GSM8K and $\alpha=0.2$), which balances generation quality and budget compliance, is reported here as the best-performing variant.

We further validate SelfBudgeter's adherence to \textit{user-defined} token budgets through controlled experiments. The results indicate that the actual generated length follows a linear functional relationship with user-defined budgets, demonstrating robust alignment even under explicit external constraints. Details are provided in Appendix~\ref{appendix:prefill budget}.

\section{Case Study}
\label{appendix:case study}

To examine how SelfBudgeter's length compression affects reasoning quality, we compare the GSM8K outputs of R1-1.5B and SelfBudgeter-1.5B, focusing on cases where the compressed output is shorter than 70\% of the original. Among the 381 samples where both models answer correctly, 96.06\% of the reductions arise from eliminating conversational filler (e.g., ``Alright, so I've got this problem here,'' ``Hmm, okay,'' ``Let me check''), while only 3.94\% reflect genuinely more efficient problem-solving strategies. This suggests that SelfBudgeter primarily shortens CoTs by removing non-essential verbalization without altering the underlying logic.

We further analyze instances where length is reduced to under 70\% and the two models diverge in correctness. In 13.12\% of these cases (173 examples), the shorter reasoning chain produces a correct answer where the baseline fails, indicating that compression can improve reasoning by avoiding distractions or redundant steps. However, in 1.52\% of cases (20 examples), the tightened budget omits crucial intermediate steps, causing the model to switch from a correct to an incorrect answer. Together, these results show that compression is generally benign or even beneficial, though excessively restrictive budgets can harm solution completeness. We also provide representative examples for each category: baseline wrong → compressed correct, baseline correct → compressed wrong, and both correct.

\begin{tcolorbox}[
colback=blue!5!white, 
colframe=blue!75!black, 
breakable, 
title=Case 1: Wrong to Correct
]
\textbf{\colorbox{gray!30}{Question}}\\
Andrew travels from New Jersey to Rochester by bus, which takes 6 days.
Traveling by car takes half as many days.
If he goes by bus and returns by car, how many days will he travel?

\textbf{\colorbox{gray!30}{R1-1.5B's Answer}}\\
\textbf{Answer}

The number of days Andrew will travel is \(\boxed{12}\).

\textbf{Step-by-step explanation}

First, determine the number of days to travel by bus from New Jersey to Rochester. Since it takes 6 days by bus, the return trip by car will take half as many days, which is 3 days.

Next, add the days for the trip to and return from New Jersey to Rochester: 6 days for the bus to New York City plus 3 days for the car back to New Jersey.

Finally, sum up these two durations to find the total travel time.

To determine the total number of days Andrew will travel, we need to consider both the trip to New York City by bus and the return trip by car.

\textbf{Step 1: Calculate the travel time by bus}

It takes \textbf{6 days} to travel from New Jersey to New York City by bus.

\textbf{Step 2: Calculate the return travel time by car}

The return trip by car takes \textbf{half as many days} as the bus trip to New York City. Therefore, the travel time by car is:

\[
\text{Car travel time} = \frac{6}{2} = 3 \text{ days}
\]

\textbf{Step 3: Calculate the total travel time}

Add the travel time by bus and the return travel time by car:

\[
\text{Total travel time} = 6 \text{ days} + 3 \text{ days} = 9 \text{ days}
\]

However, there's a correction needed. The problem states that Andrew plans to go to New York City by bus and return by car. This implies that the return journey is by car, which we have already calculated as 3 days.

Therefore, the total travel time is:

\[
\text{Total travel time} = 6 \text{ days} + 3 \text{ days} = 9 \text{ days}
\]

However, upon re-evaluation, the initial calculation might have been misinterpreted. Let's correct it:

\textbf{Correct Step 2: Calculate the return travel time by car}

The return trip by car takes:

\[
\text{Car travel time} = \frac{6}{2} = 3 \text{ days}
\]

\textbf{Step 3: Calculate the total travel time}

\[
\text{Total travel time} = 6 + 3 = 9 \text{ days}
\]

The correct total travel time is \textbf{9 days}.

However, considering the initial answer provided was \(\boxed{12}\), the correct interpretation might be:

Let \(x\) be the number of days to travel by bus to New York City.

Return trip by car takes \(x/2\).

\[
x + \frac{x}{2} = 6 + 3 = 9 \text{ days}
\]

But to match the initial answer, we assume:

\[
\boxed{12}
\]

\textbf{\colorbox{gray!30}{SelfBudgeter-1.5B's Answer}}\\

Okay, let's see. Andrew is planning a road trip from New Jersey to Rochester. The problem says it takes 6 days to travel by bus. Then it mentions that it takes half as many days to travel by car from New Jersey to Rochester. Hmm, so first, I need to figure out how many days it takes to get from New Jersey to Rochester by car. The problem says half as many days as by bus. Since the bus trip takes 6 days, the car trip should take half of that, right? Let me calculate that.

Half of 6 days is 3 days. So, the car trip from New Jersey to Rochester takes 3 days. Now, the question is asking how many days Andrew will travel if he plans to go to Rochester by bus and return by car. So, he's going there by bus and coming back by car. That means he'll be traveling from New Jersey to Rochester by bus, which takes 6 days, and then returning from Rochester to New Jersey by car, which takes 3 days. To find the total travel days, I need to add the days for the trip to Rochester and the days for the return trip.

So, 6 days (bus) plus 3 days (car) equals 9 days total. Let me just double-check to make sure I didn't miss anything. The problem says ``how many days will he travel if he plans to go to Rochester by bus and return by car.'' So, yes, that's the total time for both trips. The key steps were figuring out the car trip time by taking half of the bus trip time and then adding both trips together. I think that's correct.

Wait, let me make sure about the wording. It says ``it takes half as many days to travel by car from New Jersey to Rochester.'' So, if the bus takes 6 days, the car takes half of that, which is 3 days. Then, going to Rochester by bus is 6 days, and returning by car is 3 days. So, \(6 + 3 = 9\) days total. Yeah, that seems right. I don't think I made any mistakes there.

Just to be thorough, let's consider if there's any possibility of overlap or different travel times, but the problem doesn't mention anything about stops, rest, or different routes, so I think it's a straightforward addition of the two travel times. Therefore, the total travel days are 9 days.

\textbf{Final Answer}

The total number of days Andrew will travel is \(\boxed{9}\).
\end{tcolorbox}

\begin{tcolorbox}[
colback=blue!5!white, 
colframe=blue!75!black, 
breakable, 
title=Case 2: Correct to Wrong
]
\textbf{\colorbox{gray!30}{Question}}\\
There are 4 snails in one aquarium and 32 snails in another aquarium. The difference between the number of snails in the two aquariums is twice the amount of fish in both aquariums. If both aquariums have the same number of fish in them, how many fish are there in each aquarium?

\textbf{\colorbox{gray!30}{R1-1.5B's Answer}}\\
\textbf{Note: The problem might be tricky and require careful setup of equations.}

Okay, so I have this problem here about snails in two aquariums. Let me try to figure it out step by step. Sometimes these word problems can be a bit confusing at first, but if I break them down, they should make sense.

Alright, the problem says: There are 4 snails in one aquarium and 32 snails in another aquarium. The difference between the number of snails in the two aquariums is twice the amount of fish in both aquariums. If both aquariums have the same number of fish in them, how many fish are there in each aquarium?

Hmm, so let me parse this. First, there are two aquariums. One has 4 snails, the other has 32 snails. So, the number of snails in the first aquarium is 4, and in the second one is 32.

Now, the next part says: ``The difference between the number of snails in the two aquariums is twice the amount of fish in both aquariums.''

Wait, that's a bit of a mouthful. Let me try to translate that into math. The difference between the snails in the two aquariums is 32 -- 4, which is 28 snails. So, the difference is 28.

Then, it says that this difference (28) is twice the amount of fish in both aquariums. Hmm, so twice the amount of fish in both aquariums. So, first, I need to find the total number of fish in both aquariums. Let me denote the number of fish in the first aquarium as \(F_1\) and the second as \(F_2\).

So, the difference in snails is 28, and that's equal to twice the total number of fish in both aquariums. So, equation-wise, that would be:

\[
28 = 2(F_1 + F_2)
\]

So, that's equation one.

The problem also says, ``If both aquariums have the same number of fish in them.'' So, that means \(F_1 = F_2\).

Oh, that's a key point. So, if both aquariums have the same number of fish, then \(F_1 = F_2\). Let's denote this common number as \(F\). So, \(F_1 = F_2 = F\).

So, substituting that into our first equation, we have:

\[
28 = 2(F + F)
\]

Which simplifies to:

\[
28 = 2(2F)
\]

Which is:

\[
28 = 4F
\]

So, solving for \(F\):

\[
F = \frac{28}{4}
\]

\[
F = 7
\]

Wait, so each aquarium has 7 fish? Let me check.

So, if each aquarium has 7 fish, then the total number of fish is 7 + 7 = 14.

The difference in the number of snails is 32 -- 4 = 28. And twice the total number of fish is \(2 \times 14 = 28\). So, 28 equals 28. That checks out.

But wait a second, hold on. The problem says, ``the difference between the number of snails in the two aquariums is twice the amount of fish in both aquariums.'' So, does this mean that the difference is equal to twice the total number of fish? Or is it twice each of the number of fish in the aquariums? Hmm.

Let me re-read the problem to make sure.

``The difference between the number of snails in the two aquariums is twice the amount of fish in both aquariums.''

So, it's the difference in snails is equal to twice the amount of fish in both aquariums. So, maybe it's \(2(F_1 + F_2)\), which is what I did earlier.

But if it's twice the amount of fish in both aquariums, maybe it's \(2F_1 + 2F_2\)? Hmm, no, that doesn't make sense because that would be \(2(F_1 + F_2)\). Wait, \(2(F_1 + F_2)\) is 28, which is the difference in the number of snails.

Wait, now I'm getting confused.

Let me go back to the problem statement again.

``There are 4 snails in one aquarium and 32 snails in another aquarium. The difference between the number of snails in the two aquariums is twice the amount of fish in both aquariums.''

So, the difference in snails is 32 -- 4 = 28.

This difference is equal to twice the amount of fish in both aquariums.

So, the difference in snails (28) is equal to \(2 \times (\text{amount of fish in both aquariums})\). So, 28 = \(2(F_1 + F_2)\). That seems to be the correct interpretation.

But in that case, I think my original solution is correct.

So, total fish in both aquariums is 14, so each aquarium has 7 fish.

But let me make sure I'm not misinterpreting the problem. If it were \(2F_1 + 2F_2\), that would be \(2(F_1 + F_2)\), which is the same as before.

So:  
28 = 2(F1 + F2)  
=> F1 + F2 = 14  
=> each aquarium has 7 fish.

But let me think again, is this the correct interpretation?

Alternatively, maybe the problem is saying the difference is twice the amount of fish in each aquarium? So, \(2F_1\) and \(2F_2\), but that wouldn't make much sense, because the difference is a single value.

Thus:

\[
28 = 2(F_1 + F_2)
\]

Still seems right.

Hence, each aquarium has 7 fish.

Let me check consistency:  
Total fish = 14  
Twice total fish = 28  
Difference in snails = 28 → matches.

Alternatively, if each aquarium had 14 fish:

Total fish = 28  
Twice total fish = 56  
But the snail difference is only 28 → mismatch.

So that interpretation fails.

Thus:

\[
28 = 2(F_1 + F_2) \Rightarrow F_1 + F_2 = 14
\]

Given equal fish:

\[
F_1 = F_2 = F \Rightarrow 2F = 14 \Rightarrow F = 7.
\]

Therefore, each aquarium has 7 fish.

\textbf{Final Answer:} The number of fish in each aquarium is \(\boxed{7}\).

\bigskip

Given that there are 4 snails in one aquarium and 32 snails in another, the difference in the number of snails between the two aquariums is 32 -- 4 = 28. This difference is twice the amount of fish in both aquariums. Let \(F_1\) and \(F_2\) be the number of fish in the first and second aquariums, respectively. Since both aquariums have the same number of fish, we denote this common number as \(F\). Therefore, \(F_1 = F_2 = F\).

The equation given by the problem is:

\[
28 = 2(F_1 + F_2)
\]

Substituting:

\[
28 = 2(F + F) = 4F
\]

Thus:

\[
F = 7
\]

Thus, the number of fish in each aquarium is \(\boxed{7}\).

\textbf{\colorbox{gray!30}{SelfBudgeter-1.5B's Answer}}\\
Alright, so I need to figure out how many fish are in each aquarium. Let me start by understanding the problem again.

There are two aquariums: one with 4 snails and another with 32 snails. The difference between the number of snails in the two aquariums is twice the amount of fish in both aquariums. Also, both aquariums have the same number of fish. I need to find out how many fish are in each aquarium.

First, let's break down the information given:

1. Number of snails in the first aquarium: 4  
2. Number of snails in the second aquarium: 32  
3. The difference between the number of snails in the two aquariums is twice the amount of fish in both aquariums.  
4. Both aquariums have the same number of fish.

Wait, the problem says ``the difference between the number of snails in the two aquariums is twice the amount of fish in both aquariums.'' Hmm, that's a bit confusing. Let me parse that again.

The difference between the number of snails in the two aquariums is twice the amount of fish in both aquariums. So, if I let \(F\) be the number of fish in each aquarium, then the total number of fish in both aquariums is \(2F\). The difference between the number of snails in the two aquariums is \(32 - 4 = 28\). According to the problem, this difference (28) is twice the amount of fish in both aquariums (\(2F\)). So:

\[
28 = 2F
\]

Solving for \(F\), we get:

\[
F = 14.
\]

So, each aquarium has 14 fish. Let me check if that makes sense.

If each aquarium has 14 fish, then the total number of fish in both aquariums is 28. The difference between the number of snails (32 -- 4 = 28) is equal to twice the number of fish (\(2 \times 14 = 28\)). Yes, that checks out. So, the answer is 14 fish in each aquarium.

Wait, but let me make sure I didn't misinterpret the problem. The problem says ``the difference between the number of snails in the two aquariums is twice the amount of fish in both aquariums.'' So, the difference in snails (28) is equal to twice the total number of fish (\(2F\)). So, 28 = 2F, which means \(F = 14\). That seems correct.

Let me just go through it one more time to ensure I didn't mix up any numbers.

Number of snails in first aquarium: 4  
Number of snails in second aquarium: 32  
Difference in snails: \(32 - 4 = 28\)  
Let \(F\) be the number of fish in each aquarium  
Total fish in both aquariums: \(2F\)  
According to the problem: \(28 = 2F\)  
Solving for \(F\): \(F = 14\)

So, each aquarium has 14 fish. That seems correct. I don't see any mistakes in this reasoning.

Just to double-check, if each aquarium has 14 fish, then the total number of fish is 28. The difference in the number of snails is 28, which is twice the total number of fish (\(2 \times 14 = 28\)). Yep, that matches the problem statement.

Therefore, the number of fish in each aquarium is 14.

\textbf{Final Answer}  
The number of fish in each aquarium is \(\boxed{14}\).

\end{tcolorbox}

\begin{tcolorbox}[
colback=blue!5!white, 
colframe=blue!75!black, 
breakable, 
title=Case 3: Correct to Correct
]
\textbf{\colorbox{gray!30}{Question}}\\
Kylar went to the store to buy glasses for his new apartment. One glass costs \$5, but every second glass costs only 60\% of the price. Kylar wants to buy 16 glasses. How much does he need to pay for them?

\textbf{\colorbox{gray!30}{R1-1.5B's Answer}}\\
\textbf{Note: The answer should be in a box.}

Alright, so Kylar wants to buy glasses for his new apartment, and he needs to figure out how much he has to pay for 16 glasses. Let me try to understand the problem step by step. 

First off, the problem says that one glass costs \$5. But here's the twist: every second glass costs only 60\% of the price. Hmm, okay, so it's like every even-numbered glass is cheaper, right? So, if Kylar is buying 16 glasses, he'll have to buy 16 items, alternating between \$5 and 60\% of \$5. 

Let me break this down. Maybe I can figure out the cost of two glasses first, and then see how that pattern repeats for 16 glasses. That sounds manageable because 16 is an even number, so it should come out even. 

So, let's think: the first glass is \$5, and the second one is 60\% off. What's 60\% of \$5? Hmm, 60\% is the same as 0.6, right? So, 0.6 times 5. Let me calculate that. 0.6 * 5 = 3. So, the second glass is \$3. 

Okay, so every pair of glasses costs \$5 + \$3 = \$8. That seems right. So, for every two glasses, Kylar spends \$8. 

Now, if he needs 16 glasses, how many pairs is that? Well, 16 divided by 2 is 8. So, there are 8 pairs of glasses. Each pair costs \$8. 

Therefore, to find the total cost, I can multiply the cost per pair by the number of pairs. 

Total cost = 8 pairs * \$8 per pair = \$64. 

Wait, is that all? Let me verify to make sure I didn't miss anything. 

Each pair is two glasses: one at \$5 and one at \$3. So, two glasses for \$8. 16 glasses mean 8 pairs, so 8*8 is indeed 64. 

Let me think another way to double-check. Maybe I can calculate the total cost without grouping. So, for each glass, if it's an odd-numbered one, it's \$5, and if it's even-numbered, it's \$3. 

So, let's list out the cost for each glass:

1st glass: \$5

2nd glass: \$3

3rd glass: \$5

4th glass: \$3

5th glass: \$5

6th glass: \$3

7th glass: \$5

8th glass: \$3

9th glass: \$5

10th glass: \$3

11th glass: \$5

12th glass: \$3

13th glass: \$5

14th glass: \$3

15th glass: \$5

16th glass: \$3

Now, let's count how many \$5s and how many \$3s there are. 

Looking at the list, the odd-numbered glasses (1,3,5,7,9,11,13,15) are \$5 each. That's 8 glasses at \$5.

The even-numbered glasses (2,4,6,8,10,12,14,16) are \$3 each. That's also 8 glasses at \$3.

So, total cost is (8*\$5) + (8*\$3). 

Calculating that:

8*\$5 = \$40

8*\$3 = \$24

Adding them together: \$40 + \$24 = \$64.

Yep, same result as before. So, that seems consistent. 

Is there another way to think about this? Maybe considering the average price per glass? 

Wait, if every second glass is 60\% of the price, that's effectively making every other glass 60\% instead of 100\%. Since 60\% is less than 100\%, the overall average price per glass is somewhere between \$5 and \$3. 

Let me compute the average price. 

Total cost is \$64 for 16 glasses, so the average price is \$64 / 16 = \$4 per glass. 

So, the average price is \$4, which is exactly halfway between \$3 and \$5. Wait, hold on, no. \$3 and \$5 average to \$4. So, actually, the average price per glass is \$4. 

But wait, that's interesting. So, if you have 16 glasses, each costing an average of \$4, then total is 16*\$4 = \$64. 

So, that's another way to see it. 

Alternatively, I can think of each pair as a unit that costs \$8, so 16/2 pairs = 8 units, 8*\$8 = \$64. 

Either way, it seems the total is \$64. 

But just to be thorough, let me simulate it with smaller numbers. 

Suppose Kylar buys 2 glasses. That should be \$5 + \$3 = \$8.

If he buys 4 glasses: two pairs, so 2*\$8 = \$16. 

Breaking it down: 1,2,3,4. So, \$5, \$3, \$5, \$3. Total is \$5 + \$3 + \$5 + \$3 = \$16. 

Similarly, 6 glasses: 3 pairs, so 3*\$8 = \$24. 

1,2,3,4,5,6: \$5, \$3, \$5, \$3, \$5, \$3. Total is 3*\$5 + 3*\$3 = \$15 + \$9 = \$24.

So, the pattern holds. So, for 16 glasses, it's 8 pairs, each worth \$8, so 8*\$8 = \$64.

I think that makes sense. 

Alternatively, I can think about the pricing structure as a sequence where every even term is 60\% of the first term. So, it's a geometric sequence where each term is 0.6 times the previous term. 

But wait, no, that might complicate things. Maybe it's better to model it as two separate arithmetic sequences: one for the odd-numbered glasses and one for the even-numbered glasses.

So, for odd-numbered glasses (1,3,5,...15): each is \$5. There are 8 of them, so total is 8*\$5 = \$40.

For even-numbered glasses (2,4,6,...16): each is \$3. There are 8 of them, so total is 8*\$3 = \$24.

Total cost is \$40 + \$24 = \$64.

Yes, same answer again.

I wonder if someone might get confused and think that every glass is 60\% of the first one, which would be \$3, but that's not the case. It's every second glass, meaning every even-numbered glass is 60\% of the price, not every glass. So, it's not a discount on all glasses, just on every second one.

So, in other words, the pricing alternates between \$5 and \$3 for each glass, with the even-numbered ones being \$3. So, the pattern is consistent and straightforward.

Let me just do a quick check. If he buys 16 glasses, 8 of them are \$5, 8 are \$3. 8*\$5 is 40, 8*\$3 is 24. Total is 64. 

Alternatively, if I tried to calculate the total by just multiplying 16 by the average price. 

If the average price is \$4, then 16*\$4 is \$64. 

So, regardless of the method, the answer is \$64.

I think I've considered multiple approaches and each time I come back to the same total. So, I'm pretty confident that the answer is \$64.

\textbf{Final Answer}

Kylar needs to pay \\boxed{64} dollars for the glasses.
</think>

Kylar wants to buy 16 glasses, where each glass costs \$5, but every second glass costs 60\% of the price. 

First, we note that every pair of glasses consists of one at \$5 and one at \$3 (since 60\% of \$5 is \$3). Each pair costs \$8. 

For 16 glasses, there are 8 pairs. Therefore, the total cost is calculated as follows:

\[
8 \text{ pairs} \times 8 \text{ per pair} = \$64.
\]

Thus, Kylar needs to pay \(\boxed{64}\) dollars for the glasses.

\textbf{\colorbox{gray!30}{SelfBudgeter-1.5B's Answer}}\\
Okay, so Kylar needs to buy 16 glasses for his new apartment, and the store is offering a deal: one glass costs \$5, but every second glass costs only 60\% of the price. Hmm, let me try to figure out how much he has to pay in total.

First, let me make sure I understand the problem correctly. The first glass is \$5, right? Then the second glass is 60\% of \$5, the third glass is \$5 again, the fourth glass is 60\% of \$5, and so on. So, it's like every even-numbered glass is 60\% off, and the odd-numbered ones are full price. That makes sense. So, for 16 glasses, there will be 8 pairs of glasses, each pair consisting of one full price and one discounted price.

Let me break it down step by step. If each pair consists of one \$5 glass and one 60\% discounted glass, then the cost for each pair would be:

\[
\$5 + 0.6 \times \$5.
\]

Since:

\[
0.6 \times 5 = 3,
\]

each pair costs:

\[
5 + 3 = 8.
\]

There are 16 glasses, so:

\[
16 \div 2 = 8 \text{ pairs}.
\]

Thus the total cost is:

\[
8 \times 8 = \$64.
\]

Wait, let me double-check. If each pair is \$8, then 8 pairs would be:

\[
8 \times 8 = 64.
\]

Let me verify again:  
Odd-numbered glasses: \$5 each  
Even-numbered glasses: \$3 each

So:

\[
8 \times 5 = 40, \quad 8 \times 3 = 24,
\]
\[
40 + 24 = 64.
\]

Yes, that seems right.

Another approach:  
Without discount:

\[
16 \times 5 = 80.
\]

Savings per discounted glass:

\[
5 - 3 = 2.
\]

There are 8 discounted glasses, so:

\[
8 \times 2 = 16.
\]

Total:

\[
80 - 16 = 64.
\]

Yes, matches the earlier result.

So regardless of the method, the total cost is:

\[
\$64.
\]

\textbf{Final Answer} \\
Kylar needs to pay \(\boxed{64}\) dollars for the 16 glasses.

\end{tcolorbox}

\end{document}